\documentclass[manuscript,screen]{acmart}
\usepackage{svg}
\usepackage{pgfplots}
\usepackage{multirow}
\usepgfplotslibrary{external}
\pgfplotsset{width=7cm,compat=1.16}
\usepackage{amsthm}
\usepackage{etoolbox}
\usepackage{caption}
\usepackage{subcaption}
\usepackage{soul}
\newcommand{\dquotes}[1]{``#1''}

\AtBeginDocument{%
  \providecommand\BibTeX{{%
    \normalfont B\kern-0.5em{\scshape i\kern-0.25em b}\kern-0.8em\TeX}}}

\setcopyright{acmlicensed}
\acmJournal{CSUR}
\acmYear{2024} \acmVolume{1} \acmNumber{1} \acmArticle{1} \acmMonth{1}\acmDOI{10.1145/3657631}

\newcommand{\argmax}{\operatornamewithlimits{argmax}}

\definecolor{blue(pigment)}{rgb}{0, 0, 0}
\def\updated#1{{\color{blue(pigment)}#1}}

\begin{document}

\title{Interactive Question Answering Systems: Literature Review}

\author{Giovanni Maria Biancofiore}
\authornote{Corresponding author}
\email{giovannimaria.biancofiore@poliba.it}
\orcid{0000-0003-1317-8695}

\author{Yashar Deldjoo}
\email{yashar.deldjoo@poliba.it}
\orcid{0000-0002-6767-358X}

\author{Tommaso Di Noia}
\email{tommaso.dinoia@poliba.it}
\orcid{0000-0002-0939-5462}

\author{Eugenio Di Sciascio}
\email{eugenio.disciascio@poliba.it}
\orcid{0000-0002-5484-9945}

\author{Fedelucio Narducci}
\email{fedelucio.narducci@poliba.it}
\orcid{0000-0002-9255-3256}

\affiliation{%
  \institution{Polytechnic University of Bari}
  \streetaddress{E. Orabona, 4}
  \city{Bari}
  \state{Italy}
  \postcode{70126}
}

\renewcommand{\shortauthors}{G.M. Biancofiore, Y. Deldjoo, T. Di Noia, E. Di Sciascio, and F. Narducci}

\begin{abstract}
Question-answering systems are recognized as popular and frequently effective means of information seeking on the web. In such systems, information seekers can receive a concise response to their queries by presenting their questions in natural language. Interactive question answering is a recently proposed and increasingly popular solution that resides at the intersection of \textit{question answering} and \textit{dialogue systems}.  On the one hand, the user can ask questions in normal language and locate the actual response to her inquiry; on the other hand, the system can prolong the question-answering session into a dialogue if there are multiple probable replies, very few, or ambiguities in the initial request. By permitting the user to ask more questions, interactive question answering enables users to interact with the system and receive more precise results dynamically.

This survey offers a detailed overview of the interactive question-answering methods that are prevalent in current literature. It begins by explaining the foundational principles of question-answering systems, hence defining new notations and taxonomies to combine all identified works inside a unified framework. The reviewed published work on interactive question-answering systems is then presented and examined in terms of its proposed methodology, evaluation approaches, and dataset/application domain. We also describe trends surrounding specific tasks and issues raised by the community, so shedding light on the future interests of scholars. Our work is further supported by a GitHub page synthesising all the major topics covered in this literature study. \href{https://sisinflab.github.io/interactive-question-answering-systems-survey/}{https://sisinflab.github.io/interactive-question-answering-systems-survey/}
\end{abstract}



\maketitle

\newcommand{\userquery}[1]{\ensuremath{\overset{u}\rightsquigarrow #1}}
\newcommand{\systemquery}[1]{\ensuremath{\overset{s}\rightsquigarrow #1}}
\newcommand{\useranswer}[1]{\ensuremath{\overset{u}\rightarrow #1}}
\newcommand{\systemanswer}[1]{\ensuremath{\overset{s}\rightarrow #1}}

\section{Introduction}
\label{sec:introduction}
\noindent \textbf{Motivation.}  In the early literature, question-answering systems (QASs) were 
frequently contrasted to Search Engines (SEs), \updated{with the primary distinction being the nature of the output - SEs offer a list of documents ranked by relevance, while QASs provide a definitive answer to user inquiries~\cite{DBLP:journals/nle/HirschmanG01}.} However, \updated{the once-clear boundary between these systems is becoming ever more indistinct}, a transformation driven in large part by breakthroughs in artificial intelligence (AI), \updated{specifically Generative AI (GenAI) and Large Language Models (LLMs), which are reshaping the landscape of digital interaction and information retrieval~\cite{Zdrok2023GenerativeQA,DBLP:journals/nle/HirschmanG01}, as of the writing of this survey. Such technologies have not only expanded the capabilities of SEs and QASs but also underscored the growing need for Interactive Question Answering (IQA) systems that can engage users in a more~\textit{dynamic} conversational exchange.} 

\updated{The confluence of SEs and QASs into a more integrated landscape is evident in modern digital assistants and online search experiences. For example, querying Google for specific information such as \dquotes{President of the United States} no longer yields a mere list of documents for manual exploration but offers a concise and direct answer extracted from the Web.} This shift towards knowledge-based, direct response systems can be attributed to the complex understanding and processing capabilities of the systems, which underpin the operation of contemporary QASs.

Essentially, QASs share the characteristic of providing a clear answer to the user inquiry, irrespective of the question's nature— factual (\dquotes{Which kingdom does the animal Bird of Paradise belong to?})~\cite{bqa22}, visual (\dquotes{What colour hair does the woman have?} combined with a picture of a woman)~\cite{iqa139}, or open-goal-oriented (\dquotes{How can I connect my Fitbit sensors to the server?})~\cite{iqa146}. \updated{The response may either be extracted directly from a segment of a document or synthesized by distilling numerous textual fragments into a unified, coherent response. The methodologies of \textit{extraction} and \textit{generation} form the foundational pillars of a question-answering algorithm's architecture.}

\updated{The shift from static, one-shot queries to \textit{multi-turn dialogues} facilitated by Interactive Question Answering (IQA) systems marks a significant move towards enhancing user interaction through conversational engagement. This trend is vividly demonstrated by the global embrace of virtual digital assistants, including Amazon Alexa and Google Assistant, which shows the seamless integration of conversational AI into daily life and decision-making processes across various sectors (health, weather, and e-commerce, etc.~\cite{DBLP:conf/ucami/Berdasco0D0G19}). Moreover, the emergence of GenAI models such as ChatGPT~\footnote{\href{https://openai.com/blog/chatgpt/}{https://openai.com/blog/chatgpt/}}, Aplaca~\cite{alpaca}, Gemini~\cite{team2023gemini}, Vicuna~\cite{vicuna2023} etc. has introduced unprecedented versatility in generating responses across a wide spectrum of queries. Today, this versatility is not limited to textual information but extends to coding assistance, educational content generation, and even creative storytelling~\cite{team2023gemini}.} In these contexts, interactions extend beyond mere question-answering to include follow-ups, clarifications, and even content generation—emphasizing the crucial role of IQA in fostering a deeper, more interactive learning and problem-solving experience.

\updated{Interactive QA systems (IQASs) significantly advance the capabilities of traditional QASs by introducing mechanisms for ongoing dialogue, which are crucial for addressing the inherent limitations of non-interactive models, particularly in the \textbf{disambiguation} and \textbf{clarification} of user queries. A primary issue in traditional QASs is their inability to handle ambiguous requests without further input effectively. For example, a query like \dquotes{When was Milan founded?} could yield results related to the Milan football club instead of the city due to the ambiguous nature of the term \dquotes{Milan.} Unlike these traditional systems, IQASs employ interactive features such as follow-up questions to narrow down the search context, thereby distinguishing between the city of Milan and the football club.}

\updated{Building on the capacity to clarify ambiguity, interactive QASs also excel in expanding upon the user's initial inquiry, adding depth and breadth to the \textbf{exploration} of topics. For instance, after resolving the ambiguity in a query about Milan’s foundation to focus on the city, the system might suggest related questions such as \dquotes{What are the major historical events in Milan?} or \dquotes{What is the cultural significance of Milan in Italy?} This capacity to suggest related topics not only ensures that the system addresses the user's immediate query but also opens avenues for further exploration, enriching the user's understanding of the subject and encouraging a deeper engagement with the content.}

\updated{Among IQASs, Conversational QA Systems (CoQAS) represent a more complex challenge, dealing with specific issues of natural language processing, such as maintaining the context of a conversation across multiple turns. This requires managing the dialogue's \dquotes{state} to handle references and continuity effectively, ensuring a coherent and relevant exchange. For example, these systems must cope with complex problems like the ellipsis phenomenon, where continuity and reference are essential (e.g., \dquotes{Where was the President of the United States born?}, followed by \dquotes{Where did \textbf{he} graduate from?}). This ability to keep track of the interaction state enhances the system's utility by accommodating a broader range of inquiries and conversational dynamics, marking a significant evolution in how machines understand and respond to human language.} Such stateful interaction enhances the system's utility by accommodating a broader range of inquiries and conversational dynamics, marking an important evolution in how machines understand and respond to human language.

\noindent \textbf{Contributions.} In this comprehensive survey, we start by providing formal definitions that describe the purposes for which Question Answering Systems (QASs) are implemented in the literature, further delineating the primary characteristics that Intelligent QASs (IQASs) ought to embody, as detailed in Section~\ref{sec:approaches}. Furthermore, we present a unified and encompassing architecture highlighting the primary components of IQASs and their operation, offering a thorough exploration in Section~\ref{sec:methods}. Our analysis also includes an in-depth classification of the state-of-the-art IQASs from both methodological and application perspectives, meticulously discussed in Sections~\ref{sec:methods} and \ref{sec:evaluations}. Lastly, we provide detailed insights into the most frequently used datasets, alongside adopted evaluation protocols and metrics, meticulously organized by the paradigms and objectives of IQASs, as comprehensively outlined in Section~\ref{sec:evaluations}.
\begin{table}[!htp]\centering
\caption{List of the most frequent and useful acronyms used in this document arranged in alphabetic order. Unless specified, the final \textit{S} in an acronym means \textit{system}.}\label{tab:acronyms}
\footnotesize
\scalebox{0.95}{
\begin{tabular}{rl|rl|rl}
\hline
\multicolumn{2}{r|}{\multirow{2}{*}{\small\textbf{General}}} & \multicolumn{2}{r|}{\multirow{2}{*}{\small\textbf{QA Categories}}} & \multicolumn{2}{r}{\multirow{2}{*}{\small\textbf{Model}}} \\ 
& & & & & \\ \hline
\textbf{Acr.} & \textbf{Meaning} & \textbf{Acr.} & \textbf{Meaning} & \textbf{Acr.} & \textbf{Meaning} \\ \hline
AI & Artificial Intelligence    & QA                            &  Question Answering     & BERT                               & Bi-directional Encoder from Transformer                           \\
CA & Conversational Agent   & CB QA                             & Classifier-based QA   & Bi-LSTM                               & Bi-directional Long-short Term Memory                         \\
FAQ & Frequently Asked Question    & CoQA                               & Conversational QA    & CNN                                & Convolutional Neural Network                                   \\
HCI & Human-Computer Interaction    & $\text{CoQA}^s$                              & System-driven Conversational QA    & DAW                               & Directional Attention Weaving                                                        \\
IE & Information Engine     & $\text{CoQA}^u$                             & User-driven Conversational QA   & GRU                               & Gated Recurrent Unit                          \\
IR & Information Retrieval  & IQA                              & Interactive Question Answering     & LSTM                               & Long-short Term Memory                                               \\
KM & Knowledge Management    & $\text{IQA}^d$                                & IQA for Disambiguation    & ResNet                               & Residual Network                                                                    \\
KS & Knowledge Source   & $\text{IQA}^e$                               & IQA for Exploration     & R-CNN                               & Region-based Convolutional NN                                                           \\
LLM & Large Language Model      & KB QA                               & Knowledge-based QA      & VGGNet                               & Visual Geometry Group Network                             \\
ML & Machine Learning       & MRC QA                               & Machine Reading Comprehension QA & &                    \\
NLP & Natural Language Processing       & cQA                               & Community QA & &                                             \\
QAM & Question Answering Module         & SDTT                               & Specific Domain-Dependent Task & &                          \\ 
SE & Search Engine      & SP                               & System Performance     & &                                                        \\
ST & State Tracker              & UIA                               & Usability and Interaction Analysis            & &                           \\
VDA & Virtual Digital Assistant         & UTTP                               & User experience, Trust, and Transparency & &                      \\ 
NN &  Neural Network  & VQA                               & Visual Question Answering & & \\ \bottomrule
\end{tabular}
}
\end{table}
\subsection{Strategy for Literature Search}
\label{subsec:literature_strategy}
To identify relevant publications for this survey, we implemented a multi-level search strategy, initially focusing on top conferences in QA and knowledge management (e.g., EMNLP, SIGIR, CIKM, KDD, ESWC, WSDM) to gather relevant research. Recognizing the importance of other venues in information retrieval and natural language processing, we expanded our search to include publications from venues like ACL, ACM, IEEE, AAAI, and Neurocomputing via Google Scholar. Additionally, free searches on Google Scholar yielded papers from various other sources, mainly prioritizing conference proceedings and journals over workshop publications. Our search utilized stemmed keywords such as "interact question answer" and "explain question answer" to find papers relevant to our survey's scope.

Subsequent to reviewing these publications, we explored their bibliographies and used Google Scholar's "Related Work" feature, generating a second list of pertinent studies from a broader range of venues. Although we likely did not capture every publication on interactive question answering, we believe we've included a significant and relevant selection. The detailed list of works reviewed is available on our GitHub for reproducibility.\footnote{\url{https://sisinflab.github.io/interactive-question-answering-systems-survey/}}

\section{Question Answering Approaches}
\label{sec:approaches}
The focus of this survey is on IQASs, which can be seen as a multidisciplinary topic combining several fields such as information retrieval (IR), natural language processing (NLP), human-computer interaction (HCI) and, more recently, artificial intelligence (AI), machine learning (ML) and knowledge management (KM). Thanks to the latest developments in NLP and ML areas, QASs have found a remarkable and growing interest from the AI community, especially for the role that they play for digital assistants (e.g., Google Assistant, Amazon Alexa, Apple Siri). QASs have usually been placed within the macro area of information retrieval systems (IRSs) because of their apparently similar function, enough to consider them as a sophisticated form of IRSs~\cite{iqa70}. Nowadays, QASs have evolved to gain their own field of study, whose goals continually increase by including new topics like knowledge representation and semantic entailment.

In this section, we present the formal definition of 
IQA, including the QA problem (cf. Def.~\ref{def:IR_QA_def}), two configurations of IQASs (cf. Def.~\ref{def:IQA_disambiguation}, Def.~\ref{def:IQA_exploration}) and CoQASs (cf. Def.~\ref{def:CoQA_def}). Then, we provide defnitions of interactive session (cf. Def.~\ref{def:interactive_session}), QA state (cf. Def.~\ref{def:qa_state}) and conversational history (cf. Def.~\ref{def:CoQA_history}), and different examples/pointers to state-of-the-art solutions that would help the reader to obtain a profound understanding of the topic.

\noindent \textbf{Assumption.} In information-access systems the user's information need could be expressed through~\textit{keywords} (retrieval), a~\textit{question} (question-answering), or~\textit{user profile} (recommender system), while answers could be a piece of text, an image, or items of interest, most relevant to the information need. Systems such as task-oriented chatbots or dialogue systems whose primary role is not providing information access remain outside the focus of this survey. 

In order to highlight the motivations behind this assumption, we consider the following definition.

\begin{definition}[QA problem] 
\label{def:IR_QA_def}

Let $Q=\{q_1,...,q_N\}$ be the set of possible queries and $A=\{a_1,...,a_M\}\cup\{\mathtt{NULL}\}$ the set of possible answers including the $\mathtt{NULL}$ symbol representing the situation where the system is not able to provide an answer. 
A QA system aims to find the most relevant answer to a given question in a single shot iteration. More formally:
\[\forall q \in Q, \ \hat{a} = \argmax_{a\in A} g(q,a)\]
with $g$ being a utility function that considers how well a given answer $a$ satisfies the proposed query. In probabilistic terms, we can define the problem as finding the most
probable answer given an input question (and its context)
\[\forall q \in Q, \ \hat{a} = \argmax_{a\in A} p(a | q)\]
\end{definition}
Users may interact with QASs to find the information they need. A QAS will answer the question with a unique result well-formed in natural language, like: \textit{"The Lord of the Rings' writer is J.R.R. Tolkien"}, saving the user to search for the needed information.

\subsection{Interactive Query Answering as Exploration and Disambiguation}
\label{sec:explore-disambiguate}
Finding the correct answer to a given question is the primary goal to be achieved by QASs. Although most of the QASs show high performance for this task, some intrinsic natural language issues cannot be solved by only analyzing the submitted question.
The definition of a well-disambiguated request through a natural language question is not a trivial task. Indeed, it requires the usage of specific terms plus a cognitive effort that is not affordable for everyone. Hazrinaa et al.~\cite{DBLP:journals/ipm/HazrinaSIMN17} examine the semantic QA task, which permits the disambiguation of queries by leveraging context or requesting further information from the user.

An extension of QASs, named \textbf{Interactive Question-Answering} systems, overcome this limitation with two specific goals:

\begin{description}
    \item[Disambiguation.]  In the case, there are \textit{too many} or \textit{too few} eligible answers for a given question, or there is ambiguity in the request, the system can ask a new question to the user~\cite{konstantinova2013interactive}. For this reason, IQASs can be seen \updated{as being} halfway between \textit{QASs} and \updated{multi-turn QASs}.
    Let us consider the case when the original query $q$ has a set $A' \subseteq A$ of candidate answers. The system can suggest a new query $q' \in Q$. The user answer $a'$ to $q'$ gives the information that leads to an unambiguous answer $\hat{a} \in A'$ to the previous question $q$. We can say that $\hat{a}$ is an answer to the combination of $q$ and $q'$.
    
    \item[Exploration.] After the system returned the answer $\hat{a}$ to the initial query $q$, a new set of queries $Q'$ can be suggested by the system or posed by the user\footnote{$Q'$ is also referred to as \textit{follow-up questions} in the literature.} in order to explore other relevant topics related to $\hat{a}$.
\end{description}


\updated{A common requirement for disambiguation and exploration is} interactivity\updated{, which} refers to the possibility of the system to pose/suggest new queries to the user. In both cases, when the system suggests a query, in principle, it might have already found one or more answers to the user request, thus resulting in starting up a disambiguation and exploratory step.
\updated{However, a noteworthy difference is} that while the disambiguation step is always \textbf{system driven}, exploration can also be \textbf{user driven}. 
In the latter case, the user decides the new aspects to explore related to $\hat{a}$.

Disambiguation and exploration may run for one step only or for a sequence of steps. Depending on the "memory" the QAS has about previous questions and answers, we may have a \textbf{stateless} QAS or a \textbf{stateful} one. This latter situation leads to what we call \textbf{Conversational Question-Answering} system.
It is worth noting that during a conversation, in \textit{system driven} interactions, we only have sequences of disambiguation or exploration steps. We do not have situations where the two steps are interleaved unless the systems allow a \textit{user-driven} interaction.

In a conversation aiming to disambiguate the original query $q$, we may have situations where the answer $a'$ to $q'$ does not lead to $\hat{a}$. In these cases, the system computes a new query $q''$ based on $a'$ and so on until $\hat{a}$ is finally returned. \updated{Therefore, it is not helpful to have exploratory steps while disambiguating $q$ to compute $\hat{a}$.}

\begin{example}[IQA for disambiguation]
\label{ex:CoQA_disambiguation}
Let us consider the case in which a user needs some information about music tracks, so she asks \textit{"Who sang the song Money"}. This question \updated{is} ambiguous since the answer could refer to both the group \textit{Pink Floyd} and the musician \textit{David Gilmour}.
In this case, the system replies with a question trying to disambiguate the user information needs. It searches relevant information from the set of possible answers to reach this goal.
As a consequence, it will pose disambiguation questions until it reaches what the user meant. In this example, the system asks \textit{"Do you mean the band who sang Money?"} and the user agrees, receiving then the final answer \textit{"Pink Floyd"}.
\end{example}

It is worth noting that in the Example \ref{ex:CoQA_disambiguation} the choice of which disambiguating question to pose the user merely depends on the implementation of the system. In fact, asking for \textit{"Do you mean the musician who sang Money?"} leads to the same answer \textit{Pink Floyd} once the user feedback is given (i.e. she disagrees). That is because Example \ref{ex:CoQA_disambiguation} has a binary mutually exclusive ambiguity. In general, approaches for choosing disambiguating questions from a set that may be large depending on whether the system allows optimized interactions.

Two scenarios are possible in an exploratory conversation, different from disambiguation. The queries may be computed by the QAS (system-driven) or the user (user-driven). In a system-driven scenario, new questions are proposed for which the answer is already known. Thus, disambiguation steps become useless.

\begin{example}[IQA for system driven exploration]
\label{ex:coQA_sd_exploration}
Following the Example \ref{ex:CoQA_disambiguation}, once the final answer is reached, a QAS could start an exploratory session within the conversation. It could propose questions like \textit{"Do you know when Pink Floyd were founded?"} or \textit{"Do you know when David Gilmour joined Pink Floyd?"} and then provide the related data depending on the user's answers. For instance, the user may know when the band was founded, but she does not \updated{know} the answer to the second question, so the system will reply: \textit{"He joined the Pink \updated{Floyd} in 1968 as a support to Syd Barrett"}.
\end{example}

On the other side, in a user driven scenario, the new queries are posed by the user, which may lead to ambiguous answers. In those cases, the system needs to start disambiguation steps to reach the relevant answer.

\begin{example}[IQA for user driven exploration]
\label{ex:coQA_ud_disambiguation_exploration}
Going back to the conversation in Example \ref{ex:coQA_sd_exploration}, the user may continue the dialogue by asking: \textit{"When was he born?"}. Here, \updated{she} can refer to both \textit{David Gilmour} and \textit{Syd Barret}, so the system will ask \textit{"Do you mean Syd Barret?"}. As a consequence, the user may agree with the system or not. In the latter case, the QAS answers with \textit{"David Gilmour was born on 6 March 1946"}. The conversation will \updated{last} until the user information need is totally satisfied.
\end{example}

\updated{In summary, although both disambiguation and exploration steps allow further interaction with users outside receiving a single answer, sometimes also incorrect, they differentiate in the final goal they target. Disambiguation is designed to extrapolate the intended meaning of user questions to reach the most relevant answer whenever questions are ambiguous. Differently, exploration pushes further the user information needs, aiming at amplifying the knowledge of the human counterpart.}

With respect to \textit{stateful} exploratory QAS it \updated{is} useful to avoid interaction loops.
In principle, the exploration of the knowledge space may run forever. In fact, in case we do not consider previous interactions and answers by users, the exploration may get stuck in a loop where the system computes questions already suggested in the previous steps.

Before we give a formal definition of the different interaction steps we have discussed so far, we introduce two unary operators to represent the possible actions an IQAS or a user may perform, e.g. generating a new query or an answer. We use $y \in \{u,s\}$ to state if the next query or the answer has been generated by the user ($u$) or by the system ($s$). Given $x\in A\cup Q$, $y \in \{u,s\}$, $a\in A$ and $q \in Q$ we will use:

\begin{itemize}
    \item $x \overset{y}\rightsquigarrow q$: the IQAS ($y = s$) or the user ($y=u$) generates a new query.
    \item $x \xrightarrow{y} a$:  the IQAS ($y = s$) or the user ($y=u$) generates an answer.
\end{itemize}

Given a query $q$, we then represent the simple query-answering step as $q \systemanswer{\hat{a}} $. Analogously, for the exploratory IQA, we have $q \systemanswer{\hat{a}} \systemquery{q'}$ while the disambiguating Interactive QA steps are represented with $q \systemquery{q'} \useranswer{a'} \systemanswer{\hat{a}}$. In this case, $a'$ is the answer (e.g. feedback) provided by the user to the query $q'$ generated by the system. 

\begin{definition} [Interactive Question Answering for Disambiguation] 
\label{def:IQA_disambiguation}
An interactive question answering for disambiguation $\mathbf{IQA^d}$ is a system that takes a user query $q$ as input and computes a new query $q'$ to reach the answer $\hat{a}$ to $q$. More formally, we have
\[\ \hat{a} = \argmax_{a\in A, q'\in Q} {\textit{p}} \  (q' | q), {\textit{p}} \  (a | q \systemquery{q'} \useranswer{a'})
\]
\end{definition}


The idea here is to find the best next question $q'$ given the initial query $q$ in order to compute the best answer to $q \systemquery{q'} \useranswer{a}$. As for ${\textit{p}} \  (q' | q)$ and ${\textit{p}} \  (a | q \systemquery{q'} \useranswer{a'})$, in principle, we do not make any assumption on their (in)dependency. \updated{However, different $q'$, which lead to separated disambiguating interactive steps, could be equally probable to maximize the final answer probability independently of the actions taken to disambiguate the user query (i.e., they all lead to the same final answer). In such a case, Definition~\ref{def:IQA_disambiguation} can be further declined by separately optimizing the two probabilities, s.t.:
\[ \hat{a} = \argmax_{q'\in Q} {\textit{p}} \  (q' | q) \cdot \argmax_{a\in A} {\textit{p}} \  (a | q \systemquery{q'} \useranswer{a'})
\]
}

\begin{figure}[h!]
    \centering
    \includegraphics[scale=0.4]{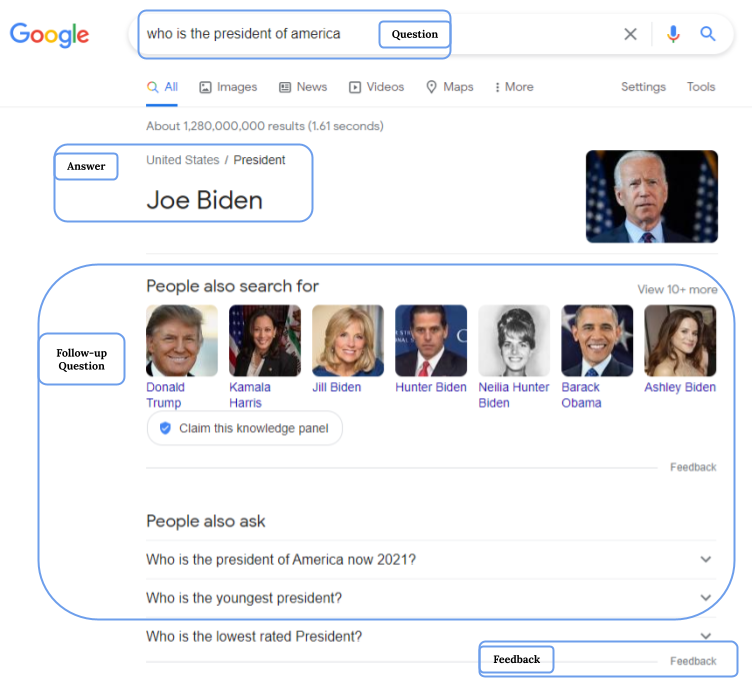}
    \caption{Example of an interactive QAs answer.}
    \label{fig:google}
\end{figure}

Figure \ref{fig:google} depicts an example of IQAS.
The question asked to the interactive system is: \textit{"Who is the president of America?"}. As already highlighted in the example, the user obtains an answer that satisfies her query plus some more suggestions useful to explore the related exploration space. Here, the picture outlines 
a collection of information and interactive area that allows users to continue their exploratory interaction.

The first information shown by Google in the example is the answer \textit{"Joe Biden"}, which includes the entities and relations of a knowledge graph recognized in the question, respectively \textit{"United States"} and \textit{"President"}. The second one proposes a set of follow-up queries that could come after the initial question. For instance, other people searched for \textit{"Who was the previous president of America?"} that was \textit{"Donald Trump"}, or \textit{"Who is the vice-president of America?"} that is \textit{"Kamala Harris"}. In this way, the user may reach all this data by simply interacting with the suggested images, enriching the solution to her information need.
The same area also groups different clickable questions asked by people to achieve the same above-correlated answers. In a way, it gives information about the context understood by the system. The user who obtains an unexpected answer could rephrase her question to disambiguate its sense intention/meaning and then receive the appropriate response from the system.

Furthermore, the system provides a suite of possible multimodal interactions, like images, audio and text, which aim at improving both the user experience and the correctness of the results.
A feedback slot allows people to interact further with the system by offering comments about the answer. Whether the solution is wrong, the user can specify the motivations and send them to the system, which will improve its behaviour about that type of question.

\begin{definition}[Interactive Query Answering for Exploration]
\label{def:IQA_exploration}
An interactive question-answering system for exploration $\mathbf{IQA^e}$ aims at guiding the user through the exploration of a knowledge space. Given a question $q$, it provides both an answer and a set of follow-up questions \updated{close} to $q$. Formally, we have:
\[\ \langle \hat{a},\hat{q}' \rangle = \argmax_{q'\in Q, a \in A} {\textit{p}} \  (a|q), {\textit{p}} \  (q'|a)
\]
\end{definition}

The aim of an $IQA^e$ is to find an answer to $q$ and, at the same time, suggest a new query $q'$ that is related to the computed answer. Without loss of generality, in the previous definition we consider only one follow-up question $q'$. The definition can be easily extended to a set $Q'$ of follow-up questions. Also in this case, we do not make any assumption on the (in)dependency between $p(q'|a)$ and $p(a|q)$. This is something that \updated{could be implemented within} the $IQA^e$.

Therefore, we may surely also have a hybrid situation where a disambiguation step supports the exploration: 
\[
q \systemquery{q'} \useranswer{a'} \systemanswer{\hat{a}}\systemquery{q''}
\]
where the initial query is disambiguated before the answer $\hat{a}$ and the next question $q''$ are computed. In a nutshell, an IQAS allows the user to further interact with the system once its reply is given via \textbf{interactive sessions}.

\begin{definition} [Interactive Session]
\label{def:interactive_session}
A set of user-enabled actions following the system reply to a previous user question defines an \textit{interactive session}:
\[
I = \{a',q' : a \overset{u}\rightsquigarrow q', q \overset{u}\rightarrow a'\}
\]
\end{definition}

The \textit{interactive session} is always supported by the IQAS to users. The cardinality of this set determines which actor leads the interaction, while its elements define the interaction type (i.e. exploratory or disambiguating). 
An \textit{interactive session} with infinite cardinality remarks the interaction led by the user. This is the case of some IQASs for exploration where users are free to pose any question to the system response. In contrast, a finite \textit{interactive session} cardinality outlines interactions conducted by the system. Therefore, a bounded number of actions (i.e. answers to disambiguating questions or enabled exploratory questions) are proposed by the IQAS to the user. Here, the interaction takes place when the user gives her feedback to the \textit{interactive session} (e.g. answering with disambiguating information).

\subsection{Conversational Query Answering}
\label{sec:conversational}
The interactive systems shown in Figure \ref{fig:google} do not keep any memory of previous interactions
with the user. When the user explores the knowledge space, every single exploration step does not consider previous ones. However, the system needs a set of information to understand the user \textit{intents} behind each question. With \textbf{context} we refer to that data helping the system in understanding the meaning intended by the user behind her questions.
Thus, the context is the set of data $C$ which supports the QAS in finding the answer $a$ to the user question $q$. Nevertheless, the overall process can be seen as a \textit{stateless} application of a sequence of $IQA^e$.
Then, the \textit{state} can be formalized as follows:

\begin{definition} [QA State]
\label{def:qa_state}
Let $q$ be the user question, $C$ a \textit{context} supporting $q$,
$a$ the system answer and $I$ the following \textit{interactive session} enabled to the user.
The \textit{QA state} is the tuple that collects all the information of a QA interaction:
\[
S_{QA} = <q, C, a, I>
\]

\end{definition}

In other words, the \textit{QA state} hosts all the data that are exchanged through a user-system interaction. Thus, it is totally outlined at the end of each interaction.

The QA state context $C$ determines the \textit{statefulness} property of an IQAS. In fact, \textit{stateless} IQASs host supporting information in their \textit{QA state context} $C$ that does not belong to other \textit{QA states}.
This can also lead to loops in the exploratory task. On the other hand, in case the system had memory of previous QA states through the \textit{context}, it could avoid proposing questions whose answers have already been visited in the past by the user. Analogously, an $IQA^d$ can only consider the original question $q$ and the answer given by the user to $q'$. In general, $IQA^d$ and $IQA^e$ are \textit{stateless} and do not take into account the \textbf{history} referring to previous interactions between the user and the system.

In case the IQA model had access to the search history of the user, it could grant new rounds of interaction helpful, e.g., in handling some linguistic issues like the \textit{Coreference Resolution} problem. The \textit{coreference resolution} is the task of finding all expressions that refer to the same entity in a text~\cite{clark2016improving}, which frequently appears by exchanging follow-up questions with the system. For example, assuming that the first user query is \textit{"Who wrote the Lord of \updated{the} Rings?"}, a follow-up question could be \textit{"When \updated{did} \textbf{he} \updated{write} \textbf{it}?"}, where \textit{"he"} refers to the answer \textit{"J.R.R. Tolkien"} and \textit{"it"} to the entity \textit{"Lord of the Rings"}. In these situations, the system must know what the user asked in the past and the given answers to solve this issue, thus moving from a \textit{stateless} configuration to a \textit{stateful} one. The \textit{stateful} configuration of an IQAS is also known in the literature as \textit{conversational} QA system due to its ability to treat conversations as having dialogues with the user. 
 
In CoQASs, we can have both user-driven and system-driven exploratory interactions as well as their combination. We will see later that, in these systems, disambiguation interactions occur as intermediate steps in a user-driven exploration. The previous example related to \textit{"The Lord of the Rings"} is a clear user-driven exploratory interaction. Indeed, thanks to the \textit{coreference resolution} and the corresponding \textit{conversational context} (c.f. Section \ref{sec:methods}) computed by the CoQAS, the user is allowed to explore the knowledge space related to their original query $q$. Analogously, a CoQAS can be used in a system-driven disambiguation process in case the answer to $q'$ is not satisfactory to provide an answer to $q$. It is worth noticing that in a \textit{system driven} CoQAS for \textit{exploration} we will never need any disambiguation step. In fact, the new queries are suggested by the system which already knows the answers (cf. Example \ref{ex:coQA_sd_exploration}). This cannot be the case for \textit{user-driven exploration} via a CoQAS. Here, a new query posed by the user after some exploration steps may require a disambiguation process by the system, (cf. Example \ref{ex:coQA_ud_disambiguation_exploration}). 
 
In a \textbf{System Driven Conversational Question Answering System for Exploration}, we have a sequence of exploratory steps in the form:
\[
q \systemanswer{a} \systemquery{q'} \systemanswer{a'} \systemquery{\ldots} \systemanswer{a^n}
\]
where the transitions between the \textit{exploratory} questions posed by the system and the related answers are interleaved by user feedback, in case she is 
knowledgeable or not on those topics.

Dually, in a \textbf{User Driven Conversational Question Answering System for Exploration} we obtain:
\[
q \systemanswer{a} \userquery{q'} \systemanswer{a'}\userquery{\ldots}\systemanswer{a^n}
\]

Differently from $IQA^e$, in the \textit{conversational} case the system-generated query $q^i$ at the $i$-th step also considers the two sets $Q^i$ and $A^i$ containing the previously generated queries $Q^i=\{q',\ldots, q^{i-1}\}$ as well as their corresponding answers $A^i=\{a,a',\ldots,a^{i-1}\}$. The same happens in the disambiguation case for the system-generated answer $a^i$.
\begin{definition}[Conversation History and Conversation Span]
\label{def:CoQA_history}
Given a conversational sequence of exploratory or disambiguation steps we define \textbf{Conversation History} at step $i$ as $h^i = Q^i \cup A^i$ where $Q^i=\{q',\ldots, q^{i-1}\}$ and $A^i=\{a,a',\ldots,a^{i-1}\}$.
For a step $l$ and a step $i$, with $l<i$ we define \textbf{Conversation Span} as
$s^{l,i} = h^i - h^l$.
\end{definition}
A conversation history contains all the user-driven and system-driven interactions up to a certain point of the conversation. They can be interpreted as the context for the next question to answer.
A conversation span represents the conversation history from a certain step $l$ up to a step $i$. We can see that $h^i = s^{0,i}$. So, what we will say for $s^{l,i}$ always holds also for $h^i$. 
Once again, we see that in a system-driven conversational exploration, at each step the system always generates the answer $a^i$ to the query $q^i$ and the next question to ask $q^{i+1}$. Also in this case, since the next query is generated by the system, it is always unambiguous. Hence, no disambiguation interaction is needed. 
This could not be the case for a user-driven conversational exploration. In fact, the user-generated query $q^i$ may result \updated{in ambiguity} to the system and then require further interaction for disambiguation. Here, we can generalize with respect to the definition of a $IQA^d$ and assume multiple steps of interaction to disambiguate $q^i$ in a situation like in the following: 
\[
q \systemanswer{a} \underbrace{\userquery{q'} \systemanswer{a'} \userquery{\ldots} \userquery{q^i}}_{exploration} \underbrace{\systemquery{\tilde{q}'} \useranswer{\tilde{a}'} \systemquery{\tilde{q}''} \useranswer{\ldots} \userquery{\tilde{q}^k} \useranswer{\tilde{a}^k}}_{disambiguation} \underbrace{\systemanswer{a^i} \userquery{q^{i+1}} \systemanswer{\ldots}\systemanswer{a^n}}_{exploration}
\]

\noindent where the conversational system is not able to disambiguate $q^i$ given a conversational span $s^{l,i}$, and then it starts $k$ multiple rounds of disambiguation steps until it gets enough information to compute the answer $a^i$. Then, the user-driven interaction may continue to explore the addressed information space. 

For the sake of presentation, we assume the \updated{system} does not have a bounded number $m$ of queries to pose during the disambiguation phase. In case it can ask at most $m$ queries to disambiguate, and after the $m$-th answer, it is not able to compute an unambiguous answer, the system returns $\mathtt{NULL}$, and the overall conversation ends.
\begin{definition} [Conversational Question Answering]
\label{def:CoQA_def}
A Conversational Question Answering system aims at exploring an information space alternatively via user-driven or system-driven interactions. 

A \textbf{system-driven Conversational Question Answering system $CoQA^s$} computes answers to the current query and the next question to ask, given a Conversation Span. More formally, we have:
\[\ \langle \hat{a},\hat{q}^{i+1} \rangle = \argmax_{q^i\in Q, a^i \in A}  {\textit{p}} \  (q^{i+1}|a^i,s^{l,i+i}),  {\textit{p}} \  (a^i|q^i,s^{l,i})
\]

In a \textbf{user driven Conversational Question Answering system $CoQA^u$} we formally distinguish between the exploration and the disambiguation as:
\[\ \hat{a} = \argmax_{a^i \in A}   {\textit{p}} \  (a^i|q^i,s^{l,i}) \quad \textrm{and} \quad \hat{q} = \argmax_{q^i \in Q}   {\textit{p}} \  (q^i|s^{l,i})
\]
\end{definition}
Please note that in $CoQA^s$ the aim is to maximize both the probability that a certain answer satisfies the current query and the probability of the next question to pose. Conversely, in $CoQA^u$, we are only interested in computing the answer, maximizing the probability that it satisfies the current query during exploration. On the other side, in a disambiguation step we are only interested in computing the next query to pose to the user in order to provide them an answer. In the following, we summarize the principal characteristics of all types of IQAS based on five dimensions that are usually found in the literature: \textit{interactivity}, \textit{statefulness}, \textit{robustness}, \textit{naturalness}, and \textit{initiative}.
\begin{description}
    \item[Statefulness] is related to the capability of the system to keep track of the state of the interaction. 
    Conversations are the most common method to obtain information and knowledge between two o more agents~\cite{qac69}. In a real-world dialogue, during the conversation, there are highly informative contextual relationships which bring new knowledge about a given topic to the interacting agents. As shown in the following example related to a $CoQA^u$, each conversation defines its dialogue context. A CoQAS stores the set of data derivable from the conversation until the end of the interaction with the user.
\begin{table}[h!]
    \centering
    \caption{An example of stateful Question-Answering Interaction}
\begin{tabular}{r l}
\toprule
\small{$q$} & \small{\textit{Who is the founder of Apple?}}\\
\hline
\small{$a$} & \small{\textit{Apple has three founders: Steve Jobs, Ronald Wayne and Steve Wozniak.}}\\
\hline
\small{$q'$} & \small{\textit{When was he born?}}\\
\hline
\small{$\tilde{q}'$} & \small{\textit{Who do you refer to?}}\\
\hline
\small{$\tilde{a}'$} & \small{\textit{Steve Jobs}}\\
\hline
\small{$a'$} & \small{\textit{24 February 1955}}\\
\hline
\small{$q''$} & \small{\textit{When was it founded?}}\\
\hline
\small{$a''$} & \small{\textit{1 April 1976, Los Altos, California, United States}}\\
\hline
\small{$q'''$} & \small{\textit{Which was the first launched product?}}\\
\hline
\small{$a'''$} & \small{\textit{The company's first product is the Apple I, a computer designed and hand-built entirely by Wozniak.}}\\
\bottomrule
\end{tabular}
\end{table}

   The first question and the related answer provide a context about Apple and its founders, enabling users \updated{to} ask for data related to both the industry and the people. In the example, the system links the \textit{"it"} pronoun of the second question with Apple, understanding that the user wants to know when Apple was founded. Moreover, thanks to the dialogue context, the CoQAS can get the implied subject of the third question, which is still Apple. In the previous example, we consider a Conversation Span $s^{0,i} = h^i - h^0$ since the information needed to answer the first question refers to the original question $q$.
   In a \textit{stateless} IQAS, every information exchanged instead is not stored by the system, so the users need to ask detailed questions each time they seek new information, resulting in less natural interactions with respect to the ones offered by \textit{stateful} IQAS.
   \item[Interactivity] According to Definitions \ref{def:IQA_disambiguation}, \ref{def:IQA_exploration} and \ref{def:CoQA_def}, IQASs and CoQASs are the only ones capable of interacting with the user. Here, the \textit{interactivity} is intended as the capability of the systems to grant users the possibility to perform varied interactions as a response to an action by the system. We call the set of user reactions made available by the system as \textit{interactive session.} In the example reported in Figure \ref{fig:google}, we can observe that after the user question, the system gives the answer but also provides an \textit{interactive session} as a set of next possible questions the user can interact with. In the case the user clicks on one of the follow-up questions listed by the system, it will react by providing the related answer and starting a new \textit{interactive session}. The interactivity is more straightly perceived in a CoQAS. In that case, a system message (reaction) corresponds to each user action.
   \item[Robustness] QASs may allow a more or less complex interaction with the user depending on a third dimension that emerges from the literature, that is \textit{robustness}. When the user question is not properly formulated or leads to ambiguous answers, a classical QAS reveals a point of failure, returning a wrong answer or $\mathtt{NULL}$. 
   This also applies to IQA$^e$ systems, designed for exploration purposes (cf. Definition \ref{def:IQA_exploration}). Conversely, IQA$^d$ and CoQA$^{s/u}$ systems allow the user to fix errors and ambiguities producing follow-up queries. IQA$^d$ systems need different \textit{interactive sessions} to solve the user errors or ambiguities due to their \textit{stateless} configuration. Instead, a \textit{stateful} setting, i.e. CoQA$^{s/u}$ systems, stores all the \textit{interactive sessions} in a context having relevant information to solve these issues from the first moment they appear. It follows that CoQASs have a higher \textit{robustness} than IQA$^d$ systems.
   \item[Naturalness]  is strictly dependant to the \textit{interactivity} of a QAS. In our analysis, a QAS is \textit{Natural} when it grants users spontaneous and immediate interactions using natural language whenever possible. It is worth noticing that naturalness is not a synonym of \textit{user friendliness} because we can have an interface that is very friendly and effective (as shown in Figure \ref{fig:google}) that is not \textit{natural} in our sense. In fact, the \textit{Naturalness} can be achieved when users cannot sense the changing of interactive sessions during the usage of a QAS. That is the case of CoQASs, where users feel a single flow of interaction instead of a sequence of separated interactive sessions. Conversely, IQASs may break this continuity, e.g., in a \textit{stateless} configuration, by posing exploratory/disambiguating questions regarding information that is already encountered by the user in the previous interactive sessions.
    \item[Initiative] Basically, the \textit{initiative} is intended as the possibility for an actor (either the user or the system) to choose which information composes the \textit{interactive session}. In detail, whether only the QAS forms the \textit{interactive session} with a limited number of admissible data the user may interact with (i.e. a set of pre-computed follow-up question or feedback to a disambiguating question), then its \textit{initiative} is \textit{system-based}. Instead, the QAS initiative is \textit{user-based} when she is free to form their own \textit{interactive session}, e.g. asking any follow-up questions to the system answer. Otherwise, we have a \textit{mixed initiative} QAS when both the previous mechanisms are enabled.
\end{description}

\subsection{Task and Challenge-based classification of Question Answering Systems}
\label{subsec:task}
QASs can be built in several \updated{ways}, covering features like robustness or naturalness and ensuring more o less sophisticated interactivity. Nevertheless, all these systems are implemented to fulfil some specific tasks, which decline the QA problem (cf. Definition \ref{def:IR_QA_def}) in multiple variations. It emerges from the literature the following \textit{task-based} taxonomy of QASs.

\begin{description}
    \item[Open-Goal QA.] Here the QASs exploit unstructured text to solve the QA problem. Forum messages or bounded sets of answers related to a specific domain, commonly known as Frequently Asked Questions (FAQ), fed the knowledge source of \textit{open-goal} QASs. More in-depth, depending on the QA model and its knowledge source, the open-goal QA is further classified into \textit{community} QA (cQA)~\cite{iqa170} and \textit{classifier-based} QA (CB QA)~\cite{iqa144}. Selecting the best answer from a pool of candidate ones, usually built on top of the forum thread, is referred to as a \textit{community} Question-Answering. In detail, cQA models foresee ranking mechanisms to achieve their goal. Conversely, a \textit{classifier-based} QAS chooses appropriate answers by categorizing questions into default classes provided by the knowledge source (e.g. FAQ). In fact, each group is mapped to a specific answer which best satisfies the related information need.
    \item[Factoid QA.] It answers questions that refer to a specific fact, e.g. \textit{"Who is Leonardo Di Caprio?"} or \textit{"What is Interstellar?"} and \textit{"Where \updated{was} Christopher Nolan born?"}. The fact answering a given natural language question has to be retrieved from the QA knowledge source, which takes the shape of a \textit{knowledge base}. The latter can occur as a set of unstructured documents hosting facts (i.e. Wikipedia, business documents, \updated{etc.}.) or as a collection of structured rules expressed in several forms (e.g. logic programming rules with Prolog and graph triples for knowledge graphs). \updated{Hence}, we distinguish \textit{machine reading comprehension} QA (MRC QA)~\cite{iqa145} from \textit{knowledge-based} QA (KB QA)~\cite{iqa152} tasks respectively. Teaching machines to read and understand texts on which to infer answers to user questions defines the \textit{machine reading comprehension (MRC)} Question-Answering task. MRC QASs reply to questions either by pointing to words span in documents or by generating a new text string, both enclosing facts satisfying the user information need. Differently, KB QAS implements a model to translate the user questions into queries allowed by the KB for seeking answers. In other words, it provides a universally accessible natural language interface to factual knowledge~\cite{iqa122}.
    \item[Visual QA.] The goal of this task is to generate answers that encapsulate a truthful description of a picture on which questions are asked. The aim of Visual QA (VQA) is to find out a correct answer for a given question which is consistent with the visual content of a given image~\cite{iqa142}.
\end{description}

\begin{table}[h!]
\caption{Work distribution over the dimensions identified.}
\scalebox{0.80}{
\begin{tabular}{c | c | c | c | c | c}
\hline
\multicolumn{1}{c|}{}& \multicolumn{2}{c|}{\textbf{Open-Goal QA}} & \multicolumn{2}{c|}{\textbf{Factoid QA}} &
\textbf{Visual QA} \\
\cline{2-6}
\multicolumn{1}{c|}{\multirow{-2}{*}{\textbf{Challenges}}} & \multicolumn{1}{c}{\textbf{cQA}} & \multicolumn{1}{c|}{\textbf{CB QA}} & \multicolumn{1}{c}{\textbf{MRC QA}} & \multicolumn{1}{c|}{\textbf{KB QA}} & \textbf{VQA} \\
\hline

\multirow{13}{*}{\textbf{SP}} & \small{Wu et al. \cite{iqa170}} & \small{Waltinger et al. \cite{iqa90}} & \small{Han et al. \cite{iqa146}}, \small{Yang et al. \cite{bqa14}} & \small{Zheng et al. \cite{iqa152}}, & \small{Shi et al. \cite{iqa161}}\\

& \small{Hu et al. \cite{iqa121}} & \small{Nie et al. \cite{eqa11}} & \small{Yuan et al. \cite{iqa145}}, \small{Chada \cite{bqa11}} & \small{Zhang et al. \cite{iqa147}} & \small{Do et al. \cite{iqa142}} \\

& \small{Wu et al. \cite{iqa117}} & \small{Su et al. \cite{inli26}} & \small{Das et al. \cite{iqa140}}, \small{Mass et al. \cite{bqa10}} & \small{Zhang et al. \cite{iqa119}} & \small{Gao et al. \cite{iqa141}} \\

& \small{Xiong et al. \cite{qac18}} & & \small{Li et al. \cite{iqa113}}, \small{McCarley \cite{bqa7}} & \small{Petukhova et al. \cite{iqa103}} & \small{Shao et al. \cite{iqa139}} \\

& & & \small{Xie \cite{iqa112}}, \small{Li et al. \cite{qac69}} & \small{Perera et al. \cite{iqa102}} & \small{Pradhan et al. \cite{iqa126}} \\

& & & \small{Bhattacharjee et al. \cite{bqa29}} & \small{Moon et al. \cite{eqa15}} & \small{Gordon et al. \cite{iqa125}} \\

& & & \small{Kuo et al. \cite{bqa25}}, \small{Kundu et al. \cite{qac66}} & \small{Damljanovic et al. \cite{inli14}} & \small{Yang et al. \cite{bqa28}} \\

& & & \small{Osama et al. \cite{bqa24}}, \small{Qu et al. \cite{qac61}} & \small{Liu et al. \cite{bqa22}} & \\

& & & \small{Wang et al. \cite{bqa21}}, \small{Su et al. \cite{qac52}} & \small{Christmann et al. \cite{qac48}} & \\

& & & \small{Qi et al. \cite{bqa20}}, \small{Li et al. \cite{qac51}} & \small{Müller et al. \cite{qac46}} & \\

& & & \small{Yang et al. \cite{bqa19}}, \small{Qu et al. \cite{qac47}} & \small{Shen et al. \cite{qac45}} & \\

& & & \small{Qu et al. \cite{bqa17}}, \small{Ju et al. \cite{qac36}} & \small{Guo et al. \cite{qac29}} & \\

& & & \small{Zhu et al. \cite{qac23}} & & \\
\hline

\multirow{5}{*}{\textbf{UTTP}} & \small{Rücklé et al. \cite{iqa115}} & \small{Liu et al. \cite{iqa98}} & \small{Chiang et al. \cite{qac68}} & \small{Sorokin et al. \cite{iqa122}} & \small{Alipour et al. \cite{iqa164}} \\

& \small{Zhang et al. \cite{eqa12}} & \small{Latcinnik et al. \cite{eqa19}} & \small{Baheti et al. \cite{qac67}}, \small{Mandya et al. \cite{qac43}} & \small{Wu et al. \cite{kq0}} & \small{Jin et al. \cite{iqa137}}\\

& \small{Zhou et al. \cite{qac22}} & \small{Sugiyama et al. \cite{qac19}} & \small{Mandya et al. \cite{qac65}} & \small{Habibi et al. \cite{qac20}} & \small{Shin et al. \cite{iqa124}} \\

& \small{Xiong et al. \cite{qac18}} & & \small{Vakulenko et al. \cite{qac59}} & \small{Xu et al. \cite{qac8}} & \small{Li et al. \cite{eqa7}} \\

& \small{Wong et al. \cite{qacon8}} & & \small{Reddy et al. \cite{qac54}}, \small{Basu \cite{qac41}} & & \small{Li et al. \cite{eqa6}} \\
\hline

\multirow{2}{*}{\textbf{UIA}} & \small{Sen et al. \cite{bqa26}} & \small{Siblini et al. \cite{qac34}} & \small{Hulburd \cite{bqa27}}, \small{Otegi et al. \cite{qac64}} & \small{Le Berre et al. \cite{bqa30}} & \multirow{2}{*}{-}\\

& \small{Kulkarni et al. \cite{qac32}} & & \small{Aken et al. \cite{bqa23}} & \small{Aken et al. \cite{bqa23}} & \\
\hline

\multirow{5}{*}{\textbf{SDDT}} & \small{Zhang et al. \cite{iqa131}} & \small{Maitra et al. \cite{iqa158}}, & \small{Schwarzer et al. \cite{iqa107}} & \small{Li et al. \cite{inli20}} & \small{Riley et al. \cite{eqa17}}\\

& \small{Kulkarni et al. \cite{qac32}} & \small{Lockett et al. \cite{iqa144}} & \small{Siblini et al. \cite{qac34}} & \small{Habibi et al. \cite{qac20}} & \\

& & \small{Lee et al. \cite{iqa138}}, \small{He et al. \cite{qac53}} & \small{Kumar et al. \cite{qac26}} & & \\

& & \small{Alloatti et al. \cite{bqa18}} & & & \\
& & \small{Sakata et al. \cite{bqa16}} & & & \\
\bottomrule
\end{tabular}}
\label{tab:works_over_sec2}
\end{table}

All the collected publications are distributed among the previously described task-oriented classes as in Table \ref{tab:works_over_sec2}.
It is noteworthy that, although a great number of works aim to realize an IQAS able to read and understand a huge number of textual documents for providing answers to users' questions, all analyzed research efforts are still evenly distributed among the highlighted categories, showing a high interest from the QA community to all these tasks.

In addition, QA approaches vary depending on the challenges they target. We found different features in QA methods aiming for different goals, e.g. optimising system performances (i.e. answers accuracy, response time, etc.) or the overall user experience. Therefore, QASs can be further classified based on the objectives they cover.

\begin{description}
\item[System Performances (SP).] 
Starting from other state-of-the-art systems, new approaches are continually proposed in order to design QASs with improved effectiveness. To this purpose, extensive offline experiments are usually carried out on several public datasets.
\item[User Experience, Trust, and Transparency (UTTP).] QASs aiming to solve this goal are designed to engage users in a more compliant way. Three main aspects are usually evaluated regarding users relationship with the system, that is \textit{user experience}, \textit{trustworthiness}, and \textit{transparency}. Whether the first one deals with \textit{satisfaction} and \textit{usability}, the \textit{trustworthiness} is more oriented to measure user expectations about results returned by the system, that is how users trust system functionalities~\cite{qac65}. The \textit{transparency} instead, evaluates the interpretability and comprehensibility of system processes~\cite{nothdurft2016justification}. 
\item[Usability and Interaction Analysis (UIA).] This challenge does not need any kind of new QAS implementation. In fact, its scope is to analyze the usage of already existing QASs or users' behaviours towards interacting with them. Thus, statistical analysis is always provided and discussed for this goal~\cite{bqa30}.
\item[Specific Domain-Dependent Task (SDDT).]
Most of papers challenging this task design a complete QA environment aiming to solve a specific issue. The proposed system usually drives users to execute domain-dependent processes or tasks in an easier way. As a consequence, these solutions strictly fit the problem for which they are designed and tested. Thus, they are not thought to be generalized. This kind of \updated{publication} often lacks comparisons with other state-of-the-art models~\cite{inli25}.
\end{description}

Table \ref{tab:works_over_sec2} also depicts the distribution of publications among the aforementioned challenges, \updated{where an unbalanced QA community interest emerges}.
Research efforts are almost totally focused on improving QAS performances, while very few works emphasize statistical analysis \updated{of} usability and interaction.
Implementation modalities and algorithms still have shortcomings in reaching strong performances about not specialized QASs, that is the IQA field is far to be solved. To underline this evidence we see a lower interest in User Experience, Trust, and Transparency challenges.

\subsection{Interaction Modalities}
\label{subsec:InteractionModalities}
IQASs allow different interaction modalities to engage users in the process of seeking answers.
Here, the interaction modality is intended as the physical channel available for users to exchange information with the system, e.g. clicking on dedicated sections of a web page, typing text messages or talking with an agent. Thus, QASs can be differentiated based on the interaction modalities they allow.
\begin{description}
    \item[Clicks \& Touches.] Here, users interact with the system by means of specific visual regions, which are designed to host only supported information. This is the case of the IQAS depicted in Figure \ref{fig:google}, where the user can only react to an interactive session by clicking on exploratory entities and questions.
    \item[Text.] The majority of QASs provide an interaction modality that foresees natural language texts as input. Users express their questions through texts and receive answers in the same format. For example, Zhang et al.~\cite{iqa131} propose a QAS which exploits a multi-scale deep neural network to extract deep semantic information from medical text to provide textual answers at different levels of granularity. That is, this framework allows only textual questions from the interaction with the user.
    \item[Speech.] This interaction modality permits users to pose their questions through speech. QASs that take advantage of users' voices are often implemented within VDAs like Siri, Google Assistant, Amazon Alexa, etc. Other relevant works are proposed by Lockett et al.~\cite{iqa144} and Mészàros et al.~\cite{inli24}. They both implement IQASs, which allow users to talk with them. The first solution addresses the operator routing task within a customer care service, where the answer is the most suitable operator to a raised issue. The latter builds a smart cart that leads users, during their shopping, to the place where an asked product is located.
    \item[Visual.] QASs allowing users to exploit images for the QA task belong to \textit{visual} QASs. This interaction modality allows visual input like images, photos or even videos. Alipour et al.~\cite{iqa164} implement an IQAS that supports images paired with natural language questions generally asking for what is represented in the picture (e.g., a photo of a football player and a question like \textit{"What sport is this?"}). Then, it computes the answer and its motivations by selecting informative \textit{region of interest} maps on the input images and exploiting \textit{deep learning} models. More straightforwardly, Shin et al.~\cite{iqa124} \updated{allow} the user to send pictures to their system, \updated{automatically answering} the question \updated{:} \textit{"What is in this picture?"}. In addition, their system asks the user several disambiguating questions to generate an answer (i.e. description) which best fits her subjective view.
\end{description}

Depending on the allowed interaction modalities, an IQAS can be further distinguished into \textit{uni-modal} (i.e., only a single interaction modality is allowed) or \textit{multi-modal} (i.e., more interaction modalities are supported) QAS. The work\updated{~\cite{sundar2022multimodal} provides a good frame of reference to the subject of multi-modality in this context.}

\section{Interactive QA Systems: Architecture and Techniques}
\label{sec:methods}
In this section we propose a general architecture of IQAS representing the prominent methodologies that exist in the literature. We also describe the implementations of each architectural element according to the state-of-the-art solutions.
\subsection{Architecture}
\label{subsec:arch}
\begin{figure}[h]
    \centering
    \includegraphics[scale=0.30]{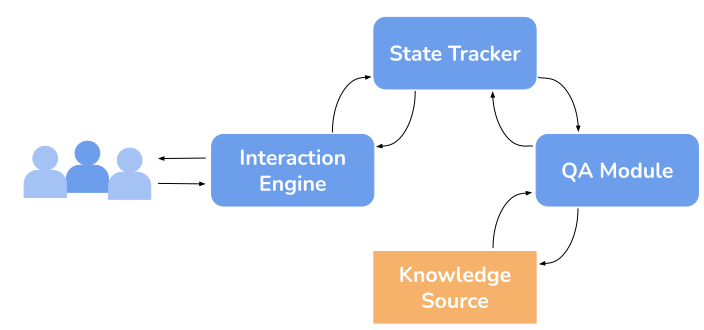}
    \caption{General Architecture of Interactive Question Answering Systems}
    \label{fig:IQAArch}
\end{figure}

Figure \ref{fig:IQAArch} outlines a high-level architecture with the components that an IQAS should implement. Four major components are proposed: the \textit{interaction engine}, the \textit{state tracker}, the \textit{QA module}, and the \textit{knowledge source}. 

The \textbf{Interaction Engine} (IE) is the module that manages the interaction with the user, receiving her requests and providing the response to her.
This module enables the interaction between the user and the system by leveraging different interfaces according to the system's capabilities. 
For instance, in case the user-system interaction is based on click (e.g. automaTA by Lee et al. \cite{iqa138}), the IE enables the system to capture the user request and prepare it in a suitable form for the \textit{state tracker} (cf. Section \ref{subsec:state_tracker}). In a similar way, when the interaction is based on natural language messages, then the IE supports the reception of textual messages \cite{iqa117}, as well as images \cite{eqa17} or even videos \cite{iqa137} for visual user-systems interactions. In a nutshell, the IE manages different \textit{interaction modalities} (cf. Section \ref{subsec:InteractionModalities}) according to the system implementation.

The complexity of the IE depends on several factors: (i) the \textit{interactivity} level, (ii) the \textit{operation modality}.
The \textit{interactivity} level refers to the distinction between standard and interactive QASs.
The IE enables users to engage with the system in both \textit{single-step} (e.g., QAS that returns the most relevant answer as a direct result to the user question \cite{bqa16}) or \textit{multi-turn} interaction (e.g., systems allow further steps to refine or expand the given system response \cite{inli26, iqa124}).
The higher the interactivity level is, the more complex the IE-module implementation will be.
The \textit{operation modality} is defined as the capability of the IE to deal with one or more \textit{interaction modes}, making an IQAS \textit{uni-modal} \cite{bqa20} and/or \textit{multi-modal}  \cite{iqa141}.
Finally, the \textit{initiative} defines whether the user \cite{qac66}, the system \cite{qac8} or both \cite{qac9} can trigger a disambiguation/exploration step, i.e. with disambiguation/follow-up questions.

The \textbf{State Tracker} (ST) manages all the information the \textit{interaction engine} and \textit{QA module} need in order to 
handle the QA request. In greater detail, it aims to fill up the \textit{QA state} (cf. Definition \ref{def:qa_state}) and to collect all the information exchanged between user and system up to that time.
In addition, referring to the Definitions \ref{def:IQA_disambiguation}, \ref{def:IQA_exploration} and \ref{def:CoQA_def}, an IQA task can also be seen as a \textit{chain} of \textit{interactive sessions} (cf. Definition \ref{def:interactive_session}) grouped by the user's information need.
Accordingly, the ST keeps track of the exact point of the \textit{chain} in which the system is at any given time: this is the \textit{QA state} that represents a sort of snapshot of the system at time $t$.

The ST operation mainly depends on two elements: (i) the \textit{context} and (ii) the \textit{tracking methods}. For CoQASs the same action could have different reactions and meanings given the \textit{dialogue context}, while for \textit{stateless} IQASs, the ST has no information about past interactions, dealing with \textit{session context} as exploitable information (e.g., images or textual snippets) that completes the user input to the system.

With regard to \textit{tracking methods}, 
in the literature they are classified as \textit{implicit} or \textit{explicit}. An \textit{explicit tracking method} keeps information about the previous \textit{QA states} in a structured form. For example, natural language tokens or knowledge graph entities can be collected \cite{kq0}.
Conversely, \textit{implicit tracking methods} store previous \textit{QA states} in a latent shape like embeddings or trained parameters of a model. That is the case of systems that rely on deep learning models, which learn patterns on real dialogue sets and perform conversations without needing explicit state tracking techniques \cite{qac61}. In that case, the state is embedded in the learned model.

The \textbf{Question Answering Module} (QAM) aims to provide the most suitable answer to the user question. This component is not specific to IQASs, but is implemented in \updated{all} QASs. It retrieves or generates the answer to the user by exploiting the context, if any, managed by the ST.
The answer depends on the \textit{data representation} adopted by the QAS (e.g., text, images, etc.) as well as the interaction mode.

In order to deal with them, the QA models can implement \textit{data-driven} and \textit{instruction-based} approaches. The former belongs to the ML field and encodes all the collected documents through \textit{numerical representations} such as TF-IDF, one-hot encoding, Word2Vec, etc. \cite{eqa7}. The latter operates  with both \textit{categorical information}, such as pure text or KG entities \cite{inli15}, and \textit{numerical representations} outlined before \cite{bqa22}.

Finally, the \textbf{Knowledge Source} (KS) stores all the knowledge exploited by QAS to answer users questions. The KS data can be \textit{structured} (e.g. Knowledge Graphs, Relational Database, etc.) or \textit{raw} (i.e. images, videos, texts, etc.). The data type can depend on the task the QAS aims to solve. For instance, a \textit{factoid} QASs can rely on both \textit{structured} information as a KG, a set of triples in the form of subject-predicate-object \cite{qac45}, and \textit{raw format} data as a collection of textual documents \cite{bqa19}. 

In the following, we aim to give an in-depth snapshot of the state-of-the-art about IQASs with respect to the components outlined in Figure \ref{fig:IQAArch}. On this line, Table \ref{tab:main_work} groups some relevant examples provided by the literature according to the IQAS modules, their aspects, and technical specifications for their implementation.
\begin{table*}[!th]
\caption{A summary of components in the IQAs representation pipeline, as sketched in Fig.\ref{fig:IQAArch}}
\centering
\resizebox{0.6\textwidth}{!}{%
\begin{tabular}{c|l l}
\hline
\multicolumn{1}{|c}{\textbf{\small{Module}}} &\multicolumn{1}{l}{\textbf{\small{Aspect}}}  &\multicolumn{1}{l|}{\textbf{\small{Approach and Example work}}} \\ \hline

\multirow{8}{*}{\small{\textbf{\begin{tabular}[c]{@{}c@{}}Interaction \\ Engine \end{tabular}}}}
& \small{Interactivity}  &  \small{single-step \cite{bqa30,iqa125},}\\
& & \small{multi-round \cite{qac62,iqa152}.} \\
& \small{Operation Modality}  &  \\
& \small{\, \, \textbullet \, Single} & \small{text \cite{qac69}, audio \cite{iqa144}.}  \\
& \small{\, \, \textbullet \, Multiple} & \small{images \& text \cite{iqa164,iqa161}, audio \& text \cite{bqa25},} \\
& & \small{video \& text \cite{bqa28}.} \\
& \small{Initiative}  & \small{system-driven \cite{iqa150,iqa147},} \\
&  & \small{user-driven \cite{iqa142,bqa17}, mixed \cite{inli14}.} \\
\hline

\multirow{2}{*}{\small{\textbf{\begin{tabular}[c]{@{}c@{}}State \\ Tracker \end{tabular}}}} 
& \small{Tracking Methods} & \small{implicit \cite{qac52}, explicit \cite{qac48,inli25}.}\\
& \small{Context}  &  \small{session \cite{eqa15}, dialogue \cite{qac61}.}\\
\hline

\multirow{11}{*}{\small{\textbf{\begin{tabular}[c]{@{}c@{}}Question \\ Answering \end{tabular}}}} & \small{Image Representation}  & \small{ResNet \cite{iqa121}, Faster-RCNN \cite{iqa139},}\\
& & \small{VGG Net \cite{iqa126}, CNN \cite{iqa125}.}\\
& \small{Text Representation}  & \small{inverted index \cite{iqa101}, pure text \cite{iqa95},}\\
& & \small{term frequency \cite{eqa6}, Word2Vec \cite{iqa121},} \\
& & \small{Glove \cite{iqa139}, LSTM \cite{iqa125}, WordPiece \cite{bqa16},} \\
& & \small{Transformers~\cite{park2023relation}, \small{GPT-3}~\cite{shao2023prompting}.}\\
& \small{Model}  &  \\
& \small{\, \, \textbullet \, Instruction-based} & \small{keyword matching \cite{iqa101,qac5},} \\
& & \small{rules execution \cite{qac48,iqa152}.} \\
& \small{\, \, \textbullet \, Data driven} & \small{supervised \cite{iqa170,iqa113}, reinforced \cite{iqa145,iqa140},} \\
& & \small{generative~\cite{tan2023can, shao2023prompting}.}\\
\hline

\multirow{2}{*}{\small{\textbf{\begin{tabular}[c]{@{}c@{}}Knowledge \\ Source \end{tabular}}}} & \small{Structured}  & \small{knowledge graph \cite{iqa122}, relational database \cite{inli15}.} \\
& \small{Unstructured}  &  \small{texts \cite{bqa23}, images \cite{eqa7}, video \cite{iqa137}.} \\
\bottomrule
\end{tabular}%
}
\label{tab:main_work}
\end{table*}

\subsection{Interaction Engine}
\label{subsec:interaction_engine}
The \textit{interaction engine} is the module that manages the user-system interactions flow. Its main objective is to allow users to communicate with the system in a natural manner.
As a result, the IE makes use of interfaces that are tailored to the system's characteristics, while also properly formatting the information intended for the user. For example, CoQASs require an IE that supports dialogues via chat-box interfaces. Rather than that, IQASs that solve the VQA task require this component to provide the capability for attaching photos as an additional feature. The IE can incorporate numerous system design choices, such as allowing users to ask their own follow-up questions (i.e. exploratory questions) or choosing a query from a predefined set. In a nutshell, the IE is characterized by three primary characteristics that include: \textit{interactivity}, \textit{operation modality}, and the \textit{authorized initiative}.

The \textbf{interactivity} of a QAS determines how users engage in the process of seeking answers. In fact, this aspect denotes the distinction between conventional QASs and IQASs. Traditionally, QASs deliver immediate responses to user queries, excluding her from further discussions. In comparison, IQASs accumulate several interaction sessions that allow both users and the system to refine/enrich the initial result. In a nutshell, interactivity defines the QAS required to support interactive user sessions.

In the literature, systems that do not anticipate the use of any interactive sessions to support achieved replies are referred to as \textit{single-step} interactive systems. Such systems receive inquiries from the user and eventually supporting data (i.e. images, document extracts, audio track, etc.) in order to produce a unique and appropriate response. Thus, the system-user interaction concludes when the user receives the system response.

For example, Aken et al. \cite{bqa23} analyze \updated{how} BERT-based QASs answer user questions. Their system addresses the MRC QA task, with a focus on understanding BERT operations in retrieving the correct answer on a given question. Therefore, authors omit the system to enable \textit{interactive sessions} for the computed answer, granting only \textit{single-step} interactions with the user.

Conversely, QASs that enable \textit{interactive sessions} to the answers given to the user are named as \textit{multi-round} interactive systems. This includes the systems described by Definitions \ref{def:IQA_exploration}, \ref{def:IQA_disambiguation} and \ref{def:CoQA_def}. The \textit{interactive sessions} allow both the user and system to refine their outcomes through \textit{multi-round} communications (e.g., disambiguating the ambiguous user questions or further exploring the system responses).

For instance, CROWN, a CoQAS implemented by Kaiser et al. \cite{qac62}, exploits a supervised approach to allow the context propagation through multi-round interactions. As a \textit{stateful} IQAS, CROWN is able to understand the context left implicit by users during their conversations. The \textit{interactive sessions} are composed of the user exploratory questions to system answers. Given the question \textit{"When did Nolan make his Batman movies?"}, the CoQAs retrieves the appropriate document snippet from its \textit{knowledge source} as the answer. Then, it enables an \textit{interactive session} in which the user can ask \textit{"Who played the role of Alfred?"}. In this case, the hidden context \textit{"Batman movies directed by Nolan"} is known since the system has memory of the previous \textit{QA states}.

It is worth noticing that the IE \textit{interactivity} has no direct relations with the \textit{statefulness} property, which belongs to the \textit{State Tracker}. Thus, the \textit{interactivity} does not distinguish \textit{stateless} IQASs from \textit{stateful} ones (i.e. CoQASs). As further proof, the example in Figure \ref{fig:google} shows an IQAS that supports multi-round interactions in a \textit{stateless} configuration.

The QAS's \textbf{operation modality} specifies which kind of interaction mode the IE has to deal with (cf. Section \ref{subsec:InteractionModalities}). QASs that manage more than one physical channel as a means of interaction (i.e. text and visual or speech and clicks) exploit a \textit{multi-modal} IE. Instead, systems that allow unique \textit{interaction modalities} in communicating with users (e.g. only text \cite{qac59} or speech \cite{bqa25}) have their IE being \textit{uni-modal}.

Qi et al. \cite{bqa20} implement a QAS based on \textit{siamese networks} to learn user preferences and answer her subjective questions. They concatenate user question representation, i.e. Bi-LSTM embeddings, with her latent factor profile. Then a classifier is trained on the resulting representations to select personal answers from a multi-choice QA dataset. To this goal, the QAS IE implements an \textit{uni-modal operation modality}, which requires only textual questions.

Conversely, \textit{multi-modal} IE deals with different \textit{operation modalities} at the same time, i.e. speech and text \cite{iqa144}, visual and texts \cite{eqa6} or texts and clicks \cite{iqa138}.

\updated{Kottur et al.~\cite{kottur2018visual} address the VQA task through conversations with the user discussing the content of a given image. Specifically, they adopt a framework of Neural Models equipped with attention layers to leverage the visual and textual information forwarded within a dialogue, where co-reference resolution may happen and be managed in multiple phases. Differently,}
Gordon et al. \cite{iqa125} \updated{implement an} autonomous agent that dynamically interacts with a visual environment to reach the answer. They design an IQAS that requires input from both a textual question and a visual context. Thus, the agent may seek other visual views for disambiguating the user question or compute the answer by means of several modules, i.e. a Scanner to capture images, a Navigator to change the system view and a Manipulator.

However, the actor mainly involved in configuring \textit{interactive sessions} is designated by the IE \textbf{initiative}, which is three-folded. We refer to \textit{user-driven initiative} when users can freely choose their questions as a follow-up to the system response (i.e. \textit{exploratory question}). In contrast, the \textit{system-driven initiative} specifies the \textit{interactive session} being a finite set of data on which users may interact (e.g. picking a follow-up question from a collection computed by the system or answering \textit{disambiguating questions}). The \textit{mixed initiative} allows both the previous two cases. 

Referring to Section \ref{sec:approaches}, systems enabling the \textit{disambiguation} have at most a \textit{system-driven initiative}, while exploratory interactions can also include a \textit{user-driven initiative}. For example, the work of Su et al. \cite{inli26} implements an IQAS which allows users to ask about the composition of API calls. In addition, it supports the user to refine the system answer by adding/removing parameters to the returned API call and exploring new related ones. The system provides these parameters as the \textit{interactive session} related to the reached answer. Thus, the proposed system \textit{initiative} is \textit{system-driven}.

Zhang et al. \cite{iqa147} designed instead an equipping tool for existing KB QASs called IMPROVE-QA with a \textit{mixed initiative} IE. It provides the capability of continuously adding new concepts/entities to the equipped QA \textit{knowledge source}, improving its accuracy. To achieve this goal, the authors implement a disambiguation step where users are asked to mark wrong and correct answers from a whole posed by the system. Moreover, they are also allowed to add further correct missing entities in the \textit{interactive session} according to the \textit{why-not} problem \cite{DBLP:conf/sigmod/ChapmanJ09}. The latter corresponds to an \textit{exploratory question} that may be answered both by inferences on the \textit{knowledge source} or by the newly added information.

Finally, Google Assistant and the system implemented by Qu et al. \cite{bqa17} are CoQASs that allow multiple user-defined interactions, which make \textit{user-driven} their \textit{initiative}. In both cases, the user is free to ask any exploratory question, moving the system to seek manifold information on a topic.

\subsection{State Tracker}
\label{subsec:state_tracker}
The \textit{state tracker} is the module that deals with \textit{QA states} as described in Definition \ref{def:qa_state}. As a middle component, it supports communications between the IE and QA modules to solve the QA task. This means that the ST takes data from both of them to update the \textit{QA state} at each interaction step. It enables the QA module to exploit information (i.e. context data) in addition to the \textit{knowledge source} for computing answers to user questions. At the same time, it supports the IE to configure the authorized interactions based on data currently stored in the \textit{QA state}.

In literature, the ST strictly depends on the \textit{statefulness} property of a QAS. It assumes different behaviours based on both \textit{QA state tracking methods} and the type of \textit{context} (cf. Definition \ref{def:qa_state}). In particular, the \textit{context} may differ in two types, i.e. \textit{session context} and \textit{dialogue context}.

The \textit{session context} is a set of supporting data contained in the \textit{QA state} that can be obtained with just a single interaction with the user. Thus, it is peculiar of \textit{stateless} IQASs. The data hosted in this type of context depends on the QAS \textit{operation modality} and the way with which that information is represented. For instance, Shi et al. \cite{iqa161} design a Quaternion Block Network to solve the VQA task, which admits images in a numeric representation in its \textit{session context}. Similarly, Das et al. \cite{iqa140} expect a set of textual paragraphs in a numerical form to support the user questions.

In contrast, Pradhan et al. \cite{iqa126} implement a QAS enhanced by Social Networks to help visually impaired people in using these platforms. The solution they propose takes advantage of a \textit{session context} shaped on multiple information sources. In fact, the ST gets contextual data like images and audio for each user question from the IE. Given the question \textit{"How many dogs are in the photo?"}, the context will have a picture as supporting visual information to find the answer besides an audio source of the question, if any, to double check any typos.

The \textit{dialogue context} is instead typical of CoQASs whose data is continuously gained after each interaction with the user. All the QA states feed the context of the next one, forming a \textit{conversation} with the user. For example, IHAF is a CoQAS implemented by Perera et al. \cite{iqa102} that hosts textual data in the form of a \textit{conceptual graph} within the context of each \textit{QA state}. During the conversation, it receives from users further information that expands the \textit{contextual graph} at each interaction step. This allows the system to return more accurate answers the longer the conversation is.

A similar intuition is adopted by Fukumoto et al. \cite{iqa95} in implementing an expansion mechanism of the \textit{dialogue context} to address the problem of ambiguous questions. Once the system receives the user query, it will ask for confirmation \updated{that} it is understood before returning an answer. The system solves an MRC QA task, allowing users to ask \textit{factual questions} like \textit{"Who is a gold medalist at Olympic game?"}. In that case, the user may intend the Olympic games held in \textit{London}, \textit{Tokyo}, or any other city. In addition, it is not known which sport the gold medalist refers to, e.g. \textit{Swimming}, \textit{Judo} or \textit{Tennis}. In fact, when the CoQAS gets entities that need a \textit{disambiguation}, it will query the user with questions like \textit{"Do you mean London Olympic game?"} or \textit{"Do you refer to the gold medalist of swimming?"}. Then, all the clue words selected by the user will feed the \textit{dialogue context} of the ST until no more disambiguation is needed. At this point, the system will be able to compute the most relevant answer to the user question. 

To keep in memory all the \textit{QA states}, the ST relies on \textbf{tracking methods}. In literature, we distinguish \textit{implicit} from \textit{explicit tracking methods} regardless the type of context hosted by \textit{QA states} (i.e. session or dialogue context).

On the one hand, the \textit{implicit tracking methods} deals with the storage of \textit{QA states} in numeric forms (e.g. latent representation). For instance, Su et al. \cite{qac52} model a CoQAS as an adaptive framework based on the sequence-to-sequence model. It manages numerical representations of all the memorized \textit{QA states} in a \textit{conversation history}, which can be combined with current questions (also numerically represented) to retrieve refined answers.

Conversely, Chiang et al. \cite{qac68} implement three CoQASs with different \textit{implicit tracking methods} to test their ability to comprehend textual contents regardless of their performances on the MRC QA task. They show how ML systems like FlowQA \cite{huang2018flowqa}, BERT \cite{devlin2018bert}, and the proposed SDNet rely more on previous \textit{QA state} answers instead of any contents host in the KS. In a nutshell, answers are computed by exclusion from the previous ones instead of being logically inferred from both the implicit tracked \textit{dialogue context} and the KS.

On the other hand, the \textit{explicit tracking methods} grant direct access to explicit data stored by the system. This information type is understandable by humans, enabling the QAS to be explainable and interpretable. Here, \textit{QA states} are not recorded in a latent form, thus preserving the structure outlined in Definition \ref{def:qa_state}. The system designed by Wong et al. \cite{qac5} represents a first attempt to realize a CoQAS with an \textit{explicit tracking method} for storing \textit{QA states}. This solution expands the \textit{dialogue context} with all the keywords that arose during the conversation with the user. Therefore, the \textit{QA Model} infers appropriate answers relying on both its KS and dialogue context, which hosts some weighted elements computed according to a decay function over conversation steps.

Instead, Zheng et al. \cite{iqa152} exploit conversations to improve the performances on the \textit{factoid QA} task. The main objective is to answer user questions through a knowledge graph, enabling cost-optimized user interactions to solve ambiguities. Thus, the ST is driven by an interaction graph which schedules the best disambiguation questions to ask requiring the lowest interaction effort. The \textit{dialogue context} is built with fragments of \textit{Basic Graph Patterns}, a set of triples usually exploited in query languages like SPARQL, that holds the facts collected during the conversation with the user. Once the final node is reached on the interaction graph, the \textit{QA module} returns the final answer.

\subsection{QA module}
\label{subsec:qa_model}

The core component of a QAS is the \textit{QA Module}, whose goal is to find appropriate answers to user questions given a \textit{knowledge source} and possible \textit{contexts}. Thus, a \textit{QA module} has to be designed to understand user queries and infer the requested information from an often large collection of data.

The \textit{QAM} differs on the type of data the QAS has to deal with and the modelling strategies that fit the QA task resolution. Therefore, it is described by two characteristics, i.e. the \textit{data representation} and the \textit{modelling approach}.

The \textbf{data representation} depends on both the supported \textit{interaction modality} and its \textit{modelling approach}, which in turn counts on the information types held by the \textit{Knowledge Source} (cf. Section \ref{subsec:knowledge_source}). In fact, we distinguish \textit{categorical} data (e.g., a sequence of textual strings) from \textit{numerical} one (e.g., vectors as well as matrices). It is worth noting that the QAM \textit{data representation} is intended as the data model that best fits the requirements of a QAS, which may differ from the ones of the \textit{knowledge source}.

In a nutshell, the QAM supports a \textit{numerical data representation} when its model requires data being expressed with numeric elements to compute an answer (e.g. dealing with images/audio or deep learning architectures). For example, the work of Hu et al. \cite{iqa121} enriches the semantic information of possible answers by adding images in their knowledge source. Thus, the QAM requires a \textit{numerical data representation} to evaluate semantic similarities between queries and possible answers. \updated{Park et al.~\cite{park2023relation} have introduced a novel method to represent textual data through meta-path tokens, an embedding approach that encodes KG information based on diverse relations along meta-paths. In this way, the authors train a QA Transformer for the KB-QA task, where the designed representation includes relationship information between node pairs of a KG in a Language-Graph shared vector space, thus allowing a simple yet effective retrieval of the right responses.}

Mandya et al. \cite{qac43} focus instead on the Co-reference resolution task in a conversational setting through ML models. In particular, they highlight a set of co-reference question-answer chains in the \textit{dialogue context} searching for terms that may refer to the same entities of the user question. Their implemented model automatically achieves the goal by means of attention mechanisms, which requires data to be represented in a \textit{numerical} form. In detail, authors embed the user question in a numeric vector by concatenating its character and word embeddings, which are obtained through a trainable numeric vector and the GloVe \cite{pennington2014glove} pre-trained model respectively. 

On the other hand, Fukumoto et al. \cite{iqa95} leverage the conversational issues by means of matching keywords methods and Named Entity Recognizer. Thus, their QAM relies on data expressed with their \textit{categorical} values.

Schwarzer et al. \cite{iqa107} implement a \textit{stateless system-driven} $IQA^{d}$ in the e-government domain for the public administration of Berlin. They indexed all the \textit{knowledge source} data (i.e. services) in an Elasticsearch\footnote{\href{https://www.elastic.co/elasticsearch/}{https://www.elastic.co/elasticsearch/}} inverted index, enriching it with meta information like services popularity rankings and additional keywords (i.e. synonyms and stems from services description words). The implemented QAM enables users to find their answers operating on the inverted indexes with three different methods (i.e. keyword scoring, Elasticsearch full-text search and custom scoring), which all require a \textit{categorical representation} of data.

The \textbf{Modeling Approach} identifies the strategies adopted by the QAS to make practical and effective reaching an answer given a question. In other words, it refers to the framework that existing algorithms or new ones exploit to solve a QA task. The choice of a model relies on three factors, which are the QA task to be solved (cf. Section \ref{subsec:task}), the supported functionalities (e.g. targeted challenges or the allowed interactivity), and the available data (i.e. the \textit{knowledge source}). Nevertheless, each implementation can be classified as an \textit{instructions-based} or \textit{data-driven} model. The former takes advantage of well-designed instruction sets to its goal, while the latter tries to take out and learn patterns from a huge set of data (i.e. query-response pairs) to reply to the user questions.

Hence, the \textit{instructions-based} methods rely on a set of rules designed a priori to accomplish its task. The resulting QAM is totally unaware of data held by the \textit{knowledge source}. Thus, its implementation covers a finite number of scenarios that may emerge during the process of answering questions. The state-of-the-art QASs can be further divided into three categories: (i) \textit{keyword matching} models, (ii) \textit{pipeline execution} algorithms, and (iii) \textit{translation} models.

The \textit{keyword matching} strategy aims to find the answer to a question based on the number of matches between their key terms. For instance, the previous work of Schwarzer et al. \cite{iqa107} implements several scoring functions that retrieve the answer among the most relevant documents based on how many question keywords are found in them. Each scoring function is designed to consider a specific keyword feature by means of different weight functions.

Conversely, Wong et al. \cite{qac5} realize a first attempt at CoQAS by concatenating keywords extracted from user questions with the ones stored in the ST \textit{dialogue context} with a gradually decaying weighting function. Thus, keywords that are far in the dialogue context have less relevance than the most recent or repeated ones. The system searches for answers, giving high priority to documents with relevant matching keywords at specific moments.

The \textit{pipeline execution} models foresee a sequence of functions performed in cascade to solve the QA task. Christmann et al. \cite{qac48} take advantage of this modelling strategy to implement Convex, a \textit{factoid} CoQAS based on the Wikidata\footnote{\href{https://www.wikidata.org/}{https://www.wikidata.org/}}. The authors design a set of processing steps to be executed in the pipeline for solving the KB QA task. Starting from the user question, a (i) named entity recognition and disambiguation (NERD) system identifies Wikidata entities contained in the text. Then, a (ii) context sub-graph is built with Wikidata entities in the neighbourhood of the one recovered by (i). Each element of this sub-graph and its frontier nodes represent a potential answer to the user question. Thus, (iii) key entities are retrieved based on three relevance evaluations (i.e. relevance to the question words, to the sub-graph context, and to the knowledge source priors). These three signals are then aggregated through (iv) the Fagin’s Threshold Algorithm \cite{fagin2003optimal} and the answer is obtained from (v) a scored list of entities, which assumes the solution be in the near proximity of the key entities and the ones stored in the \textit{dialogue context}.

Furthermore, \textit{translation} approaches perform a translation of questions from a natural language to a formal one, which depends on the QAS \textit{knowledge source}. It allows users to have direct access to complex structured data collections (e.g. knowledge graphs or relational databases) without being experts in the related query languages (i.e. SPARQL and SQL). Naeem et al. \cite{inli15} implement a QAS that translates the natural language user questions into OLAP queries, while FREyA, by Damljanovic et al. \cite{inli14}, is one of the first examples of KB QASs that translates the user questions into SPARQL queries. It identifies a set of ontology concepts from the natural language question through the combination of syntactic parsing and ontology reasoning techniques or by means of string similarities evaluation, synonyms detection and user engaging. The SPARQL query is built based on the retrieved ontology concepts. 

Differently, \textit{data-driven} strategies refer to models trained on often huge data collections to learn patterns for answering user questions. These models build their own knowledge by looking at the examples provided during a training phase. Then, they take advantage of their training to solve a specific task (i.e. answering questions). The QAM \textit{data-driven} strategies found in the literature are classified in \textit{supervised} and \textit{reinforced}.

The \textit{supervised} \textit{data-driven} model relies on a labelled dataset to learn which answers are associated with a given question example. They further be divided into \textit{detecting} and \textit{generative} implementation, which depends on the method with which the answer is computed. With the \textbf{detecting} approach the QAM learns how to detect the desired response from an information set. Thus, the answer can be either found as a limited sequence of words within a document or picked from a set of answers held by the knowledge source. For example, the work of Alloatti et al. \cite{bqa18} proposes a CB QAS on the e-invoicing domain and its regulation. The authors opt for a BERT-based model with an added layer to classify user questions into specific groups, which are, in turn, labelled with the related answer. The training procedure matches with the fine-tuning of BERT in learning a one-hot encoded vector whose indexes refer to the different group of answers.

Li et al. \cite{qac69}, instead, design a conversational MRC QAS with a directional attention weaving (DAW) mechanism to extract answers from documents \updated{held} in the \textit{knowledge source}. 
They implement a model that estimates the start and the end of the answer within a document for each question. To achieve this goal in a conversational setting, they train the model on the CoQA dataset \cite{qac54} with a combination of different attention types to leverage both data coming from the knowledge source and the dialogue context. In contrast, \textit{generative} approaches learn how to build the answer by composing appropriate words to the given question. In other words, the output is generated from scratch instead of being selected among the existing ones.

Li et al. \cite{iqa113} train a CoQAS to answer the user question about information that was previously given to the system. It expects as input a list of sentences defining a queryable "story". Then, the user asks for data about that story in a conversational setting. In fact, the system also learns to pose disambiguation questions to the user when its current knowledge lacks crucial information for computing a reply. The \textit{generative data-driven} model proposed by authors is based on encoding sentences and questions via Gated Recurrent Unit (GRU). The two results are combined and decoded by another GRU to compose the answer or a disambiguation question. This framework is trained on a large corpus of conversations to learn when a disambiguation question is needed and how a sentence (i.e. question and answer) is built by means of words. \updated{Differently, Tan et al.~\cite{tan2023can} designed an experimental setting to test the abilities of the GPT LLM family in solving the KB-QA task without employing conversations for complex questions. Here, all the LLM types receive the question expressed in natural language, and the model autoregressively generates a textual snippet on which, subsequently, the authors check for the presence/absence of the answer.}

Finally, the \textit{reinforced} \textit{data-driven} models adopt the paradigm of \textit{reinforced learning} to comprehend the QA task. Hence, the QAM performs some \textit{actions} based on \textit{observations} and \textit{states}, which may \updated{result} in \textit{transitions} to other \textit{states} eventually \textit{rewarded}. In a nutshell, the training procedure states the system in attempting to solve the QA problem through an intrinsic logic. When the QAS answers correctly, it receives a reward that usually minimizes its training cost function. Otherwise, no incentives or penalties are given to the system. A clear example is provided by Gordon et al. \cite{iqa125}, where the proposed visual QAS is modelled as an agent able to explore the environment of a given picture through several actions (i.e. navigating, manipulating, scanning and detecting as well as answering). Here, the authors enable the hierarchical reinforced learning paradigm to train a high-level controller in selecting and invoking the right sub-task (i.e. the previous actions) to reach the correct answer in an efficient way. The reward is received when the system planning produces a positive outcome (i.e. the right answer).

\subsection{Knowledge source}
\label{subsec:knowledge_source}

The information needed by the system to compute answers to user questions is hosted by the \textbf{knowledge source} (KS). Data about facts, domain-specific instructions, and community opinions are stored here in different forms and modalities. This component represents the basis on which the QAS knowledge is built, defining what it is aware of and what it does not know. The \updated{held} information is made available both at running time and at an eventual training time for the QAS. This distinction determines the type of the adopted QAM modelling approach (cf. Section \ref{subsec:qa_model}), which also depends on the kind of data contained in the KS.

Although the KS module usually mirrors the main features of datasets exploited by the system to accomplish its task (cf. Section \ref{sec:evaluations}), in this section we provide a view oriented to the QA processes instead of its data. In fact, we focus on the organization of information collections at a high level of abstraction to label the state-of-the-art QASs approaching methods with data. Hence, data modalities and formats are not concerned to be discussed here. The QAS \textit{knowledge source} can be classified into \textit{structured}, \textit{unstructured}, and \textit{mixed}.

The \textbf{structured} KS manages the information composing the system knowledge by means of well-designed structures. In other words, it accepts only data that presents a standard structured organization universally recognized (e.g. knowledge graphs, relational databases). This is the case of Zheng et al. \cite{iqa152} and Zhang et al. \cite{iqa147}, where the implemented systems are designed to rely on knowledge graphs (i.e. Dbpedia \cite{mendes2012dbpedia} and Freebase \cite{bollacker2008freebase}) to retrieve the answer. Instead, the collection of question-answer string pairs (e.g. QALD-6 and WebQuestion datasets) is used only to evaluate the system's performance. Li et al. \cite{inli20} instead foresee a relational database as their CoQAS \textit{structured} KS.

In contrast, \textbf{unstructured} KSs allow data sources to lack an intrinsic structure. They cover the majority of QASs populating \updated{today's} literature, which rely on paragraph lists or image collections to reply to user queries. Wu et al. \cite{iqa117} implement a cQA that depends on question-answer pairs (i.e. string sentences) that can be retrieved from any forum website. Instead, Gao et al. \cite{iqa141} use a list of question-images pairs labelled with answers as a KS for their VQAS. \updated{Recently, generative LLMs have introduced real-time information generation through engineered prompts into knowledge synthesis. Shao et al.~\cite{shao2023prompting} developed the Prophet framework, a VQA model that selects candidate answers for image-related queries. These answers and the image caption form the KS for the next stage, where a LLM (e.g., GPT-3) determines the most relevant response.}

Finally, the \textbf{mixed} KS enables the system to rely on both structured and unstructured sources of data. An example is given by Zhang et al. \cite{eqa12}, where the implemented QASs rely both on multi-modal knowledge graphs (with images) and collections of qualified doctor advice to consultants in the form of question-answer strings to build its knowledge. Shen et al. \cite{qac45} also exploit a combination of structured and unstructured information to train their CoQAS. In this case, authors take advantage of the Wikidata KG for the semantic knowledge and the Complex Sequential Question Answering (CSQA) \cite{1801.10314} dataset to deal with dialogue.

\section{Evaluation and Dataset}
\label{sec:evaluations}
Even while all IQASs in the literature can be characterized using the unified architecture seen in Figure \ref{fig:IQAArch}, the datasets used to train and evaluate their models do not share a common set of properties. Depending on the QAS purpose, the modality type, and the interactivity setting (i.e. QA, $IQA^{e}$, $IQA^{d}$ or CoQA), the nature and characteristics of a dataset might be vastly different. Consequently, evaluation methods also end up being rather distinct.

In this section we provide guidelines about evaluation protocols and measures usually exploited in literature to test QASs. We also 
explain which datasets constitute the state-of-the-art regarding QA tasks (cf. Section \ref{subsec:task}), QAS \textit{modeling approaches} (cf. Section \ref{subsec:qa_model}) and evaluation goals deriving form the QA challenges.

\subsection{Evaluation protocols}
\label{subsec:evaluation_prot}
The evaluation phase of IQASs is crucial for determining their effectiveness in relation to specific tasks and challenges. This phase allows for the comparison of different solutions to identify the most effective ones and highlights potential development issues. Evaluation protocols, consisting of standardized actions, are essential for assessing system performance across various case studies, each focusing on particular qualities like linguistic robustness.

There are two main types of evaluation protocols: \textit{offline} and \textit{online}. \textit{Offline} evaluations use pre-compiled datasets and cross-validation to analyze the system's prediction capabilities without real-user interaction. In contrast, \textit{online} evaluations involve real users to assess the system's functionality and the relevance of its results through user studies or operational measures. For instance, Shin et al.~\cite{iqa124} employ the BLEU algorithm to evaluate the naturalness of responses in a VQAS without user interaction, whereas Riley et al.~\cite{eqa17} measure the accuracy of their VQA solution on domain-specific scenarios.

In \textit{online} evaluations, user engagement is critical for examining the system's performance in real-world scenarios. Wong et al.~\cite{qac8} assess their CoQAS through user sessions, using Likert scales to rate system utterances, while Li et al.~\cite{inli20} focus on the usability of their SDDT KB QAS by tracking the average task completion time. Such evaluations are vital for advancing IQAS development by providing insights into both system effectiveness and user satisfaction.

\begin{table}[h!]
\caption{Common datasets and evaluation metrics used in the IQAS literature. This Table represents a snapshot of the current literature. A comprehensive list can be found on our GitHub repository.}
\scalebox{0.75}{
\begin{tabular}{l l c c c c c c c c c c c c c c}
\toprule
\small{Authors} &\small{Year} &\begin{tabular}[c]{@{}c@{}}\small{Main} \\ \small{chall.}\end{tabular} &\multicolumn{11}{c}{\small{Evaluation}} & \small{Dataset} & \small{Task} \\ \cline{4-14}

& & &\small{type} &\multicolumn{10}{c}{\small{metrics}} & & \\ \cline{5-14}

& & & &\multicolumn{5}{c}{\small{offline metrics}} & &\multicolumn{4}{c}{\small{online metrics}} & & \\ \cline{5-9} \cline{11-14}

& & & &\rotatebox{90}{\small{Efficiency }} &\rotatebox{90}{\small{Error}} &\rotatebox{90}{\small{Dialogue}} &\rotatebox{90}{\small{Answer}} &\rotatebox{90}{\small{Rank}} & &\rotatebox{90}{\small{User Imp. }} &\rotatebox{90}{\small{Rank}} &\rotatebox{90}{\small{Efficacy}} &\rotatebox{90}{\small{Answer}} & & \\ \toprule

\updated{\small{Park et al.~\cite{park2023relation}}} &\updated{\small{2023}} &\updated{\small{SP}} &\updated{\small{offline}} & & & &\updated{\small{x}} & & & & & & &\updated{CommQA, OBQA, MedQA} &\updated{FQA}\\

\updated{\small{Shao et al.~\cite{shao2023prompting}}} &\updated{\small{2023}} &\updated{\small{SP}} &\updated{\small{offline}} & & & &\updated{\small{x}} & & & & & & &\updated{\small{OK-VQA, A-OKVQA, ScienceQA, TQA}} &\updated{\small{VQA}} \\

\updated{\small{Tan et al.~\cite{tan2023can}}} &\updated{\small{2023}} &\updated{\small{SP}} &\updated{\small{offline}} & & & &\updated{\small{x}} & & & & & & &\updated{\small{CWQ, QALD-9, WQ-SP, GraphQ}} &\updated{\small{FQA}} \\

\updated{\small{Li et al.~\cite{li2023few}}} &\updated{\small{2023}} &\updated{\small{SP}} &\updated{\small{offline}} & & & &\updated{\small{x}} & & & & & & & \updated{\small{WQSP, GrailQA, GQA, MetaQA}} & \updated{\small{FQA}}\\

\updated{\small{Saxena et al.~\cite{saxena2022sequence}}} &\updated{\small{2022}} &\updated{\small{SP}} &\updated{\small{offline}} & & & & &\updated{\small{x}} & & & & & &\updated{GQA, VQA 2.0} &\updated{VQA}\\

\updated{\small{Liu et al.~\cite{liu2022declaration}}} &\updated{\small{2022}} &\updated{\small{SP}} &\updated{\small{offline}} & & & &\updated{\small{x}} & & & & & & &\updated{MetaQA, CWQ, WQSP} &\updated{FQA}\\

\updated{\small{Zhong et al.~\cite{zhong2022proqa}}} &\updated{\small{2022}} &\updated{\small{SP}} &\updated{\small{offline}} & & & &\updated{\small{x}} & & & & & & &\updated{SQuAD, Quoref, NewsQA} &\updated{FQA}\\

\updated{\small{Tafjord et al.~\cite{tafjord2022entailer}}} &\updated{\small{2022}} &\updated{\small{UTTP}} &\updated{\small{online}} & & & & & & &\updated{\small{x}} & & & &\updated{OBQA, QuaRTz} &\updated{MRCQA}\\

\updated{\small{Kacupaj et al.~\cite{kacupaj2022contrastive}}} &\updated{\small{2022}} &\updated{\small{SP}} &\updated{\small{offline}} & & & & &\updated{\small{x}} & & & & & &\updated{ConvQ, ConvRef} &\updated{FQA}\\

\updated{\small{Vakulenko et al.~\cite{vakulenko2021question}}}&\updated{\small{2021}} &\updated{\small{SP}} &\updated{\small{offline}} & & & &\updated{\small{x}} &\updated{\small{x}} & & & & & &\updated{CANARD, TREC} &\updated{MRCQA}\\

\updated{\small{Park et al.~\cite{park2021bridge}}} &\updated{\small{2021}} &\updated{\small{SP}} &\updated{\small{offline}} & & & &\updated{\small{x}} & & & & & & &\updated{TGIF-QA, MSVD-QA, MSVRTT-QA} &\updated{VQA}\\

\updated{\small{Guo et al.~\cite{guo2021re}}} &\updated{\small{2021}} &\updated{\small{SP}} &\updated{\small{offline}} & & & &\updated{\small{x}} & & & & & & &\updated{COCO-QA, VQA, VQA 2.0} &\updated{VQA}\\

\updated{\small{Jia et al.~\cite{jia2021complex}}} &\updated{\small{2021}} &\updated{\small{SP}} &\updated{\small{offline}} & & & & &\updated{\small{x}} & & & & & &\updated{TimeQuestions} &\updated{FQA}\\

\small{Wu et al. \cite{iqa170}} &\small{2020} &\small{SP} &\small{offline} & & & & &\small{x} & & & & & & \small{TREC, YH, SE, WQA} &\small{cQA} \\

\small{Alipour et al. \cite{iqa164}} & \small{2020} &\small{UTTP} &\small{online} & & & & & & &\small{x} & & & & \small{VG, VQA} &\small{VQA} \\


\small{Maitra et al. \cite{iqa158}} & \small{2020} &\small{SDDT} &\small{offline} & & & &\small{x} & & & & & & & \small{FB, DL, InsD, IndD} &\small{CBQA} \\

\small{Wu et al. \cite{kq0}} & \small{2020} &\small{UTTP} &\small{offline} &\small{x} & &\small{x} &\small{x} & & & & & & &\small{SQ} &\small{FQA} \\

\small{Latcinnik et al. \cite{eqa19}} & \small{2020} &\small{UTTP} &\small{both} & & & &\small{x} & & &\small{x} & & & &\small{ComSQA, QASC} &\small{CBQA} \\

\small{Zheng et al. \cite{iqa152}} & \small{2019} &\small{SP} &\small{both} & & & &\small{x} & & & & &\small{x} & & \small{QALD-5, WQ, GQ} &\small{FQA} \\


\small{Han et al. \cite{iqa146}} & \small{2019} &\small{SP} &\small{offline} & & & &\small{x} &\small{x} & & & & & & \small{InsQA, WPQA} &\small{MRCQA} \\



\small{Do et al. \cite{iqa142}} & \small{2019} &\small{SP} &\small{offline} & & & &\small{x} & & & & & & & \small{V7W, VQA 2.0, TDIUC} &\small{VQA} \\


\small{Das et al. \cite{iqa140}} & \small{2019} &\small{SP} &\small{offline} & & & &\small{x} & & & & & & & \small{TQA, SeQA, Q-T, SQuAD} &\small{MRCQA} \\


\small{Lee et al. \cite{iqa138}} & \small{2019} &\small{SDDT} &\small{online} & & & & & & & & &\small{x} & & \small{SG} &\small{CBQA} \\

\small{Jin et al. \cite{iqa137}} & \small{2019} &\small{UTTP} &\small{offline} & &\small{x} & &\small{x} & & & & & & & \small{TGIF-QA, MSVD-QA, MSVRTT-QA} &\small{VQA} \\



\small{Zhang et al. \cite{eqa12}} & \small{2019} &\small{UTTP} &\small{offline} & & & &\small{x} &\small{x} & & & & & &\small{S-C, SG} &\small{cQA} \\

\small{Nie et al. \cite{eqa11}} & \small{2019} &\small{SP} &\small{both} &\small{x} & & & & & & &\small{x} &\small{x} & &\small{NCD, BC, LM-1B, NC, WSC-G, COPA} &\small{CBQA} \\

\small{Zhang et al. \cite{iqa131}} &\small{2018} &\small{SDDT} &\small{offline} & & & & &\small{x} & & & & & & \small{cMedQA 2.0} &\small{cQA} \\

\small{Pradhan et al. \cite{iqa126}} & \small{2018} &\small{SP} &\small{offline} & &\small{x} & & & & & & & & & \small{COCO-QA} &\small{VQA} \\


\small{Shin et al. \cite{iqa124}} & \small{2018} &\small{UTTP} &\small{both} & & & &\small{x} & & & & &\small{x} &\small{x} &\small{COCO-QA} &\small{VQA} \\

\small{Sorokin et al. \cite{iqa122}} & \small{2018} &\small{UTTP} &\small{online} & & & & & & & & & &\small{x} &\small{-} &\small{FQA} \\

\small{Hu et al. \cite{iqa121}} & \small{2018} &\small{SP} &\small{offline} & & & & &\small{x} & & & & & &\small{SG} &\small{cQA} \\


\small{Li et al. \cite{eqa7}} & \small{2018} &\small{UTTP} &\small{offline} & & &\small{x} & &\small{x} & & & & & &\small{VQA-E} &\small{VQA} \\

\small{Su et al. \cite{inli26}} & \small{2018} &\small{SP} &\small{both} & & & &\small{x} & & &\small{x} & &\small{x} & &\small{NL2API} &\small{CBQA} \\

\small{Wu et al. \cite{iqa117}} & \small{2017} &\small{SP} &\small{offline} & & & &\small{x} &\small{x} & & & & & &\small{SemEval 2015, BZ} &\small{cQA} \\

\small{Rücklé et al. \cite{iqa115}} & \small{2017} &\small{UTTP} &\small{offline} & & & &\small{x} & & & & & & &\small{InsQA, SE} &\small{cQA} \\

\small{Wu et al. \cite{iqa113}} & \small{2017} &\small{SP} &\small{offline} & &\small{x} & &\small{x} & & & & & & &\small{bAbI, IQA} &\small{MRCQA} \\

\small{Xie \cite{iqa112}} & \small{2017} &\small{SP} &\small{offline} & & & & &\small{x} & & & & & &\small{DBQA} &\small{MRCQA} \\

\small{Schwarzer et al. \cite{iqa107}} & \small{2016} &\small{SDDT} &\small{offline} & & & & &\small{x} & & & & & &\small{LeiKa} &\small{MRCQA} \\

\small{Petukhova et al. \cite{iqa103}} & \small{2015} &\small{SP} &\small{offline} & & &\small{x} &\small{x} & & & & & & &\small{EAT} &\small{FQA} \\

\small{Perera et al. \cite{iqa102}} & \small{2014} &\small{SP} &\small{offline} & & & &\small{x} & & & & & & &\small{TREC 8} &\small{FQA} \\

\small{Li et al. \cite{inli20}} & \small{2014} &\small{SDDT} &\small{online} & & & & & & &\small{x} & & &\small{x} &\small{MASD} &\small{FQA} \\

\small{Liu et al. \cite{iqa98}} & \small{2013} &\small{UTTP} &\small{both} & & & & &\small{x} & & & & &\small{x} &\small{BZ} &\small{CBQA} \\

\bottomrule
\end{tabular}}
\label{tab:data_ev_1}
\end{table}

However, both the offline and online configurations can be exploited concurrently to realize a further detailed system evaluation. For instance, Zhang et al.~\cite{iqa147} and Shin et al.~\cite{iqa124} implement this mixed strategy to evaluate their systems in a wider range of features. The former engage 300 college students to analyze the system performance changes in increasing the number of hints given by human participants, while the latter ask up to 3000 workers to evaluate response features like diversity, attractiveness and expressiveness. In fact, these test goals are hard to be simulated and automatically evaluated.

\noindent In what follows, we will classify state-of-the-art evaluation techniques based on specific characteristics shared in relation to the objectives of the tests. In particular, these aspects are recognized as the result produced by the system (for example, a single answer, a list of ranked replies, or a dialogue), which is the primary focus of the evaluation.

\begin{description}
\item[Single Answer.] Systems are assessed based on their ability to respond to user queries, with the returned responses serving as the primary metric to be evaluated. Depending on the problems, tasks, and modeling methodologies covered by the QAS, solutions can be assessed at several levels, which in turn establish assessment objectives such as \textit{correctness}, \textit{reliability}, and \textit{sensitivity}, as well the \textit{naturalness} and the \textit{expressiveness}.
\item[Ranked List] This group refers to the system capacity to retrieve pertinent resources to a given queries. Although the QA task allows a single result to return to the user, some state-of-the-art works also permit other replies returned by the system after the initial response.
This list reveals the \textit{reasoning capabilities} of the QAS, as well as which features/information are deemed important for locating appropriate responses. The examined test objectives are comparable to those of the first group, although being able to be evaluated on a deeper level (e.g., answers positions in the returned ranked list).

\item[Interaction] Rather than the solution itself, the attention is on the interaction enabled by the system in order to arrive at it. The interaction influences the quality of the system's responses. Consequently, interactions are assessed based on their  \textit{cost} and their required \textit{user efforts}, as well as their \textit{effectiveness} and \textit{efficacy}. The majority of studies focusing on this element employs online evaluation techniques. Nonetheless, offline metrics are also utilized for stateful IQASs, which may be further evaluated based on coherence, context, and naturalness.
\end{description}

Table \ref{tab:data_ev_1}
summarizes the research findings. Each entry is characterized by (i) the reference to the specific paper, (ii) the year it was published, (iii) the primary problem addressed by the proposed solution, (iv) the type of evaluation setting used to test the system, (v) the element on which test objectives are evaluated, (vi) the dataset exploited to evaluate the implemented system (which are detailed in Section \ref{subsec:dataset}), and (vii) the QA task implemented. Although the targeted elements are similar both offline and online settings, we chose to emphasize trends over the years of the two evaluation protocols by noting them individually.

\subsection{Dataset}
\label{subsec:dataset}
There are a large number of data resources in the literature that are focused on the implementation and evaluation of QAS. \updated{Specific datasets are more suitable than others for evaluating some peculiar findings depending on the QA modelling technique and the particular QA task covered by the IQAS.}

\begin{table*}[!th]
\caption{Main relevant datasets exploited in IQASs literature grouped by tasks and QAM model types.}
\centering
\resizebox{0.55\textwidth}{!}{%
\begin{tabular}{c|l l}
\hline
\multicolumn{1}{|c}{\textbf{\small{Task}}} &\multicolumn{1}{l}{\textbf{\small{Modeling Approach}}}  &\multicolumn{1}{l|}{\textbf{\small{Dataset}}} \\ \hline

\multirow{2}{*}{\small{\textbf{cQA}}} & \small{Data Driven} & \multirow{2}{*}{\small{Yahoo!, StackExchange}}\\
& \small{Instruction Based} & \\ \hline

\multirow{2}{*}{\small{\textbf{CB QA}}} & \small{Data Driven} & \multirow{2}{*}{\small{Domain Dependant Datasets}} \\
& \small{Instruction Based} &  \\ \hline

\multirow{2}{*}{\small{\textbf{MRC QA}}} & \small{Data Driven} & \small{CoQA, SQUAD, TriviaQA, SearchQA}\\
& \small{Instruction Based} & \small{CLEF, QuAC, SQUAD}\\ \hline

\multirow{2}{*}{\small{\textbf{KB QA}}} & \small{Data Driven} & \small{SQUAD, CSQA}\\
& \small{Instruction Based} & \small{WebQuesitons, QALD-N, SimpleQuestions}\\ \hline

\multirow{2}{*}{\small{\textbf{Visual QA}}} & \multirow{2}{*}{\small{Data Driven}} & \multirow{2}{*}{\small{VQA, VQA2.0, TDIUC, COCO-QA}}\\
& & \\
\bottomrule
\end{tabular}%
}
\label{tab:main_dataset}
\end{table*}

\noindent According to the literature, the most frequently used datasets for building, testing, and evaluating an IQAS are shown in the Table~\ref{tab:main_dataset}. In the interest of completeness, we present a comprehensive list of datasets found in the literature. Each is complemented with a concise explanation of its content and structure.

\begin{description}
\item[TREC] a collection of datasets published at the Text REtrieval Conference. With the TREC dataset, we refer to all data collections belonging to the homonym QA challenges available on this \href{https://trec.nist.gov/data/qamain.html}{link}. These datasets contain questions and response patterns, as well as a pool of documents returned by competing teams~\cite{wang2007jeopardy}.
\item[Yahoo! (YH)] identifies a set of information obtained from the website Yahoo! Answers hosted by \href{http://webscope.sandbox.yahoo.com}{Yahoo!}. The dataset contains 10 million question pages of Yahoo! Answers and 100.000 user queries about them~\cite{zhou2016learning}.
\item[Stack Exchange (StEx)] is a \href{https://law.stackexchange.com/}{web platform} of exchanging information among users through the question-answer modality. The dataset collects over 7 thousand questions posted on the website equipped with related answers ordered by their relevance~\cite{anderson2012discovering}.
\item[WikiQA] is realized by Yang et al.~\cite{yang2015wikiqa} to give researchers a valuable source of data for the open-domain QA task. The main feature of this collection is the proposal of the answer-triggering task, which identifies the possible absence of the correct answer among the documents associated with each question. The structure is similar to previous datasets (questions with a set of sentences ordered by their correctness), and it is available at this \href{https://www.microsoft.com/en-us/download/details.aspx?id=52419}{link}.
\item[Visual Genome (VG)] hosts data enabling to answer questions about objects depicted in images~\cite{krishna2017visual}. The data collection is proposed to solve the VQA task, collecting dense annotations of objects, attributes and relationships for each image. It is available on the \href{https://visualgenome.org/}{author's website}.
\item[Visual Question Answering (VQA)] is a dataset firstly proposed by Antol et al.~\cite{antol2015vqa} to define the VQA task. Their dataset holds for each question a related image and a set of equally appropriate answers expressed in different forms. A second version (VQA 2.0) is realized by Goyal et al.~\cite{goyal2017making} to overcome the language biases issue that characterizes the first version. In fact, the authors demonstrate that systems trained on the previous dataset version tend to ignore the visual features due to language priors. However, both datasets are available at this \href{https://visualqa.org/}{link}.
\updated{\item[Visual Dialog (VisDial)], instead, is a benchmark designed by Das et al.~\cite{das2017visual} to address the task of VQA for CoQASs. The authors collected dialogues for 140k pictures composed of 10 question-answer pairs sequences. Such conversations focus on subjects that take the scene of the related pictures, and the dialogue's actors naturally discuss the image content to include relevant linguistic figures like Co-reference Resolution through exchanging their messages. The whole collection is available on this \href{https://visualdialog.org}{website}.}
\item[Question Answering over Linked Data (QALD)] is a collection of challenges designed to test the QA task over the Linked Data. The first appearance of this type of dataset dates back to the work of Lopez et al.~\cite{lopez2013evaluating}. All the resources are available on the QALD \href{http://qald.aksw.org/}{web platform}, which hosts at this time 9 QALD challenges.
\item[WebQuestions (WQ)] is a set of question-answers pairs built by Berant et al.~\cite{berant2013semantic}. Starting from a single factoid question, they exploited the Google Suggest API to retrieve further questions related to the first one and link a set of possible answering entities to each question.
This dataset is available at this \href{https://worksheets.codalab.org/worksheets/0xba659fe363cb46e7a505c5b6a774dc8a}{link}. It also available at this \href{http://aka.ms/WebQSP}{link} the semantic parsed (SP) version of WQ (i.e. WebQuestionsSP), which contains testing values of gathering SPARQL queries and answers for each question of the original dataset~\cite{yih2016value}.
\item[\updated{GraphQuestions} (GQ)] holds more than 5000 logical form-question pairs associated with answers from the knowledge base to enable fine-grained analysis of QASs. These queries have been automatically generated by Su et al.~\cite{su2016generating} and then refined by human operators to reduce redundancy and commonness. The whole dataset is made available a the following \href{https://github.com/ysu1989/GraphQuestions}{GitHub page}.
\item[WikiPassageQA (WPQA)] is a Wikipedia-based collection specific for non-factoid answer passage retrieval produced by Cohen et al.~\cite{cohen2018wikipassageqa}. It contains thousands of questions with annotated answers queried by human workers based on Wikipedia documents. The whole collection is available at this \href{https://ciir.cs.umass.edu/downloads/wikipassageqa/}{platform}.
\updated{\item[Open Table-and-Text Question Answering (OTT-QA)], distributed by Chen et al.~\cite{chen2020open}, is a dataset that groups a set of multi-hop questions requiring aggregating information from Wikipedia tables and text to be answered. This collection assesses the capabilities of QASs not only to aggregate information over multiple structured/unstructured sources but also to perform multi-hop steps to retrieve the most valuable data for giving the correct answers. The dataset is available at this \href{https://github.com/wenhuchen/OTT-QA}{GitHub page}.}
\item[InsuranceQA] represents a data set collecting question and answer pairs from the web in the insurance domain. It was released on the following \href{https://github.com/shuzi/insuranceQA}{GitHub page} by the work of Feng et al.~\cite{feng2015applying}.
\item[QAit] is a dataset designed by Yuan et al.~\cite{iqa145} focused on procedural knowledge to answer the user questions. In fact, the answering procedure is here treated as a text game, where agents can interact with the environment (i.e. textual descriptions) to reach the answer (i.e. objects). Replies are generated as action sequences, which depend on the environment where the question is posed. The data resource is made available on this \href{https://github.com/xingdi-eric-yuan/qait_public.}{GitHub page}.
\item[Task Driven Image Understanding Challenge (TDIUC)] identifies a data collection hosting more than 1.6 million questions organized in 12 different categories. Is was built by Kafle et al.~\cite{kafle2017analysis} to overcome the unfair evaluation of different algorithm abilities (e.g., object detection, object and attribute classification, positional reasoning, counting) by grouping samples in dedicated categories. The dataset is available at this \href{https://kushalkafle.com/projects/tdiuc.html}{web page}.
\item[Visual7W] is proposed by Zhu et al.~\cite{zhu2016visual7w} by establishing a semantic link between textual descriptions and image regions by object-level grounding. They enable VQASs to return visual answers besides textual ones, organizing the dataset in a set of questions-images-multi-choice answers triples, where each answer refers to a specific image region. Authors hold their data resource at this \href{http://ai.stanford.edu/~yukez/visual7w/}{page}.
\item[TriviaQA] is produced by Joshi et al.~\cite{joshi2017triviaqa} as a MRC QA dataset with more than 650 thousand question-answer-evidence documents triples. Their goal is to generate a resource that tests systems on their potential bias about question style or content, requiring them to select the best documents that support the answers they compute. The dataset is available at this \href{http://nlp.cs.washington.edu/triviaqa/}{link}.
\item[SearchQA] is a large-scale dataset for the MRC QA task published by Dunn et al.~\cite{dunn2017searchqa}. The collection is generated starting from existing question-answer pairs and augmenting it with text snippets retrieved by Google. Their resource hosts more than 140 thousand question-answer pairs, each of them associated on average with 49.6 snippets, and it is available on the following \href{https://github.com/nyu-dl/dl4ir-searchQA}{GitHub page}.
\item[Quasar-T] is built by Dhingra et al.~\cite{dhingra2017quasar} from the software entity tags on the Stack Overflow \footnote{\href{https://stackoverflow.com/}{https://stackoverflow.com/}} website. Each dataset record includes a question, a relevant context document, a set of candidate solutions and the correct one. It is specifically designed for the MRC QA task and it is publicly available at this \href{https://github.com/bdhingra/quasar}{GitHub page}.
\item[Stanford Question Answering Dataset (SQuAD)] is an MRC dataset consisting of questions about Wikipedia articles. Answers are extracted as a text segment (i.e. span) from the passage related to each question. The first version was published by Rajpurkar et al.~\cite{DBLP:conf/emnlp/RajpurkarZLL16} to address the need for large and high-quality resources for the MRC QA task. The dataset was updated to the 2.0 version (SQuAD 2.0) by Rajpurkar et al.~\cite{DBLP:conf/acl/RajpurkarJL18} to include unanswerable questions that look similar to answerable ones. Both the dataset are available on this \href{https://rajpurkar.github.io/SQuAD-explorer/}{link}.
\item[TGIF-QA] is a collection proposed by Jang et al.~\cite{DBLP:conf/cvpr/JangSYKK17} to address the video QA task, which requires spatial-temporal reasoning from videos to answer questions correctly. They extended the Tumblr GIF dataset~\cite{DBLP:conf/cvpr/LiSCTGJL16} with more than 100 thousand query-reply pairs. The dataset is accessible on this \href{https://github.com/YunseokJANG/tgif-qa}{GitHub page}.
\item[Microsoft Research Video Description Corous (MSVD-QA)] is a dataset introduced by Chen et al.~\cite{DBLP:conf/acl/ChenD11} that consists of 120 thousand sentences summarizing actions in short video snippets, available on this \href{https://mega.nz/#!QmxFwBTK!Cs7cByu_Qo42XJOsv0DjiEDMiEm8m69h60caDYnT_PQ}{link}.
\item[MSRVTT-QA] is a large-scale video benchmark for video understanding by Xu et al.~\cite{DBLP:conf/cvpr/XuMYR16}. The collection covers the video-to-text task, with 10 thousand web video clips and 200 thousand clip-sentence pairs. Each clip is annotated with about 20 natural language sentences thanks to a human operator. This dataset is available at this \href{https://www.microsoft.com/en-us/research/publication/msr-vtt-a-large-video-description-dataset-for-bridging-video-and-language/}{link}.
\item[Chinese Medical Question Answer dataset (cMedQA)] is built by Zhang et al.~\cite{zhang2017chinese} to cover a medical question answering task in Chinese. The resource gets questions and answers from the online Chinese health community, and it is hosted on this \href{https://github.com/zhangsheng93/cMedQA}{GitHub page}. A second version (cMedQA 2.0) is produced by Zhang et al.~\cite{iqa131} by expanding and cleaning the question-answer pairs of the previous one and is available at this \href{https://github.com/zhangsheng93/cMedQA2}{GitHub page}.
\item[COCO-QA] holds question-answer pairs for the VQA task obtained through a question generation algorithm proposed by Ren et al.~\cite{DBLP:conf/nips/RenKZ15}. It converts image descriptions into QA pairs and then rejects those that appear too rarely or too often. The overall collection is hosted on the following \href{http://www.cs.toronto.edu/~mren/research/imageqa/data/cocoqa/}{website}.
\updated{\item[Outside Knowledge VQA (OK-VQA)] originated from a subsection of images of the COCO-QA dataset selected by Marino et al.~\cite{marino2019ok} to cover a VQA task that needs further knowledge by the system outside the information packaged within the dataset. The authors' intention here is to assess whether a system, given an image, can employ it as a link to the proper external knowledge for answering the user question. The dataset is available on this \href{https://okvqa.allenai.org}{website}.}
\item[Interactive Question Answering Dataset (IQUAD v1)] identifies a set of simulated environments designed by Gordon et al.~\cite{iqa125} to evaluate agents in answering questions through interactions with surrounding objects. The dataset contains more than 75 thousand multiple-choice questions, each one equipped with a unique scene configuration. Authors published their resource on this \href{https://github.com/danielgordon10/thor-iqa-cvpr-2018}{GitHub page}.
\item[SimpleQuestions (SQ)] is a large-scale dataset based on Freebase built by Bordes et al.~\cite{DBLP:journals/corr/BordesUCW15}. It holds more than 100 thousand questions written by human annotators and associated to Freebase facts. The goal is to evaluate systems, trained on complex reasoning, solving easy to answer questions. It is available on this \href{https://github.com/davidgolub/SimpleQA/tree/master/datasets/SimpleQuestions}{GitHub page}.
\item[SemEval-2015] is a challenge designed by Nakov et al.~\cite{nakov2019semeval} to leverage the answer selection for cQASs. Authors select Arabic and English question-answers collections from two relatively communities, then labelled by human operators to be available on the SemEval Task 3 challenge, available on this \href{https://alt.qcri.org/semeval2015/task3/}{link}.
\item[bAbI] dataset is realized by Weston et al.~\cite{weston2015towards} to test systems to answer questions via chaining facts, simple induction or deduction. The collection has different independent test cases that can be eventually merged to evaluate specific scenarios. The whole work is publicly available on this \href{https://research.facebook.com/downloads/babi/}{website}.
\item[IQA dataset (ibAbI)] is designed by Li et al.~\cite{iqa113} to extend the bAbI dataset by adding interactive scenarios to the QA task. They simulate three different representative scenarios of incomplete or ambiguous information, called the ambiguous actor problem, the ambiguous object problem and the missing information problem. It adopts the same standard as the bAbI dataset, and it is available at this \href{http://www.cs.toronto.edu/pub/cuty/IQAKDD2017/}{link}.
\item[Question Answering via Sentence Composition dataset (QASC)] produced by Khot et al.~\cite{khot2020qasc} is a multi-hop reasoning dataset that requires retrieving and composing facts to answer a multiple-choice question. Each entry is composed of a question, a set of answers enclosing a single correct one and a set of supporting facts on which the system must rely to respond to the question. The dataset is available at this \href{https://github.com/allenai/qasc}{link}.
\item[Commonsense QA] is designed by Talmor et al.~\cite{konstantinova2013interactive} to investigate question answering with prior knowledge. It collects a set of commonsense questions relying on concepts and complex semantic relations to be answered. The entire dataset is hosted on this \href{https://www.tau-nlp.org/commonsenseqa}{website}.
\item[OpenBookQA (OBQA)] is a dataset about elementary-level science facts built by Mihaylov et al.~\cite{mihaylov2018can}.
This work focuses on evaluating the ability of QASs to leverage specific languages and understanding related to the science domain. The collection is available at this \href{https://leaderboard.allenai.org/open_book_qa/submissions/public}{link}.
\item[AI2 Reasoning Challenge dataset (ARC)] is designed to require systems with powerful knowledge and reasoning capabilities to answer questions. Sabharwal et al.~\cite{clark2018think} divided the collection in a challenge set and an easy one. The former contains only questions answered incorrectly by the majority of QASs to stimulate the community to design improved systems. The dataset is available on this \href{https://allenai.org/data/arc}{website}.
\item[Pororo QA] is a dataset containing more than 16 thousand scene-dialogue pairs, each one with fine-grained sentences for scene description and story-related QA pairs. The resource was built by Kim et al.~\cite{kim2017deepstory} to perform video story QA by learning from a large number of cartoon videos. It is available on this \href{https://github.com/Kyung-Min/PororoQA}{GitHub page}.
\updated{\item[Audio Visual Scene-aware Dialog (AVSD)], similarly to Pororo QA, is a collection of scene-aware conversations developed by Alamri et al.~\cite{alamri2019audio} to leverage the multi-modality feature of CoQASs. The dialogues built over 11k videos of human actions are designed to test whether a CoQAS can exploit important information in the temporal dynamics of the scene and the audio. The collection is held on this \href{https://video-dialog.com}{website}.}
\item[TVQA] is a large-scale video QA dataset based on 6 popular TV shows. Questions are designed by Lei et al.~\cite{lei2018tvqa} to be compositional in nature, requiring systems to recognize relevant moments within clips and comprehend subtitles and visual concepts to respond adequately. The resource is available at this \href{http://tvqa.cs.unc.edu/}{link}.
\item[HotpotQA] by Yang et al.~\cite{yang2018hotpotqa} is a Wikipedia-based QA dataset whose questions require searching and reasoning over supporting documents to be answered. To this, sentence-level supporting facts required for reasoning are given, which enables strong supervision and explanations. The resource is available on this \href{https://hotpotqa.github.io/}{website}.
\item[Question Answering in Context (QuAC)] is a collection of 14 thousand information-seeking QA dialogues designed by Choi et al.~\cite{choi2018quac} to deal with real-world questions that are usually open-ended, unanswerable or meaningful within a dialogue context. The dataset is available on this \href{https://quac.ai/}{website}.
\item[Conversational Question Answering dataset (CoQA)] by Reddy et al.~\cite{qac54} contains 127 thousand QA pairs obtained from 8 thousand conversations about text passages of specific domains. Questions are presented in a conversational form while answers are free-form text with corresponding evidence highlighted in the passage. This frequently used collection to test CoQASs is available on the following \href{https://stanfordnlp.github.io/coqa/}{GitHub page}.
\item[Complex Sequential QA dataset (CSQA)] is designed by Saha et al.~\cite{1801.10314} to combine the task of answering factual questions through complex inference over realistic-sized KG and learning to converse through a series of coherently linked QA pairs. It contains around 200 thousand of dialogues with a total of 1.5 million turns.
The dataset is available on this \href{https://amritasaha1812.github.io/CSQA/}{web page}.
\end{description}

\section{Conclusion}
\label{sec:conclusion}

In conclusion, we have reviewed a substantial collection of interactive question-answering systems (IQASs)-related literature published during the past decade. We discovered the literature to be diverse, beginning with adopted methodologies for addressing multiple QA tasks and concluding with a vast array of diverse resources (i.e. knowledge sources and datasets) that are typically utilized to create and evaluate question-answering systems (QASs). Despite the fact that the state-of-the-art is defined by several types of QA solutions, we were able to determine the characteristics shared by the suggested systems that constitute a shared framework. To the best of our knowledge, we are the first to present a unified and comprehensive design that emphasizes the fundamental components and functions of IQASs. For each component, we have performed an in-depth categorization of the literature from the methodological and application perspectives, categorizing the works by tasks, difficulties, and interaction modes. In addition, in order to address the demands of the IQAS community, we give explicit definitions of the implementation goals and features of IQASs. To achieve this goal, we have categorized QASs based on their behaviours and enabled interaction so that the whole IQAS landscape may be characterized using these definitions and features. Then, we detailed trends regarding specific tasks and problems, demonstrating the community's keen interest in enhancing the system's performance and openness. Lastly, we have included a basic classification of evaluation approaches often used in the literature to assess QASs, together with a full list of datasets used in the evaluation process, general descriptions, and links to the authors' hosting platforms.

Despite the fact that the state-of-the-art examined in this survey comprises a large number of solutions encompassing a wide range of IQAS elements and characteristics, we believe there is still considerable potential for advancement as future challenges. In particular, based on the developing patterns from the literature analysis, the QA community is headed toward four significant problems. A first step toward the unification of conversational search and question answering is the development of new dialogue-based search and answer algorithms. This condition is a direct result of the blurring of the lines between question-answering systems and information retrieval that is becoming increasingly apparent. The second difficulty follows the never-ending efforts to enhance the AI systems' comprehension of human semantics and ontology. This line focuses mostly on the development of techniques to extract relevant data from disparate data sources i.e., \textit{multi-modality}. It pursues a third future trend that seeks more natural and multi-modal interfaces/interaction strategies to enhance the user's enjoyment of IQAS. The final objective of future research will be to compile a comprehensive dataset that can be used to evaluate all IQAS kinds explored thus far under the constraints and tasks outlined in this literature review.

\bibliographystyle{ACM-Reference-Format}
\bibliography{bibliography}

\end{document}